\documentclass[sigconf, nonacm]{aamas} 
\makeatletter
\AtBeginDocument{%
  \fancyhead[LE,RO]{}
  \fancyhead[LO,RE]{}
}
\makeatother

\usepackage{placeins}
\usepackage{booktabs}
\usepackage{hyperref}
\usepackage{enumitem}
\usepackage{algorithm}
\usepackage{algpseudocode} 
\usepackage{balance} 

\makeatletter
\gdef\@copyrightpermission{
  \begin{minipage}{0.2\columnwidth}
   \href{https://creativecommons.org/licenses/by/4.0/}{\includegraphics[width=0.90\textwidth]{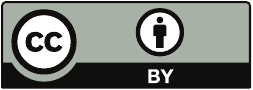}}
  \end{minipage}\hfill
  \begin{minipage}{0.8\columnwidth}
   \href{https://creativecommons.org/licenses/by/4.0/}{This work is licensed under a Creative Commons Attribution International 4.0 License.}
  \end{minipage}
  \vspace{5pt}
}
\makeatother

\makeatletter

\gdef\@acmConference{}
\gdef\@acmBooktitle{}
\gdef\@acmYear{}
\gdef\@acmDOI{}
\gdef\@acmISBN{}

\makeatother

\acmSubmissionID{2242}

\title[AAMAS-2027 Formatting Instructions]{Preference-Conditioned Multi-Objective Reinforcement Learning for Tactical Decision Making in Heavy-Duty Vehicles}

\author{Deepthi Pathare}
\affiliation{
  \institution{Chalmers University of Technology and University of Gothenburg}
  \city{Gothenburg}
  \country{Sweden}}
  \affiliation{
  \institution{Trucks Technology and Industrial Division, Volvo Group}
  \city{Gothenburg}
  \country{Sweden}}
\email{pathare@chalmers.se}

\author{Leo Laine}
\affiliation{
  \institution{Chalmers University of Technology}
  \city{Gothenburg}
  \country{Sweden}}
  \affiliation{
  \institution{Trucks Technology and Industrial Division, Volvo Group}
  \city{Gothenburg}
  \country{Sweden}}
\email{leo.laine@chalmers.se}

\author{Morteza Haghir Chehreghani}
\affiliation{
  \institution{Chalmers University of Technology and University of Gothenburg}
  \city{Gothenburg}
  \country{Sweden}}
\email{morteza.chehreghani@chalmers.se}

\begin{abstract}
A central difficulty in applying reinforcement learning to real-world applications such as autonomous driving is that conventional scalar reward formulations obtained by aggregating multiple often competing objectives may obscure the structure of their trade-offs. These challenges are particularly relevant to decision-making for heavy-duty vehicles, which requires balancing safety, efficiency, and operational costs under demanding commercial requirements. We present a preference-conditioned multi-objective reinforcement learning framework that explicitly represents these objectives and utilizes Generalized Policy Improvement and Proximal Policy Optimization to learn policies capturing different trade-offs between them. The proposed method approximates the Pareto front across three conflicting objectives: safety, quantified in terms of collisions and successful completion; energy efficiency, quantified using energy cost; and time efficiency, quantified using driver cost. We evaluate the framework in a scalable highway traffic simulation environment tailored for tactical decision making in heavy-duty trucks. It yields smooth and interpretable Pareto fronts, and evaluations across traffic densities and against an analytical baseline show that the learned policies capture meaningful trade-offs. The preference-conditioned policy enables transitions between different driving behaviors without retraining, demonstrating the potential of multi-objective reinforcement learning for adaptive decision making in autonomous heavy-duty vehicles.
\end{abstract}

\keywords{Multi-Objective Reinforcement Learning, Autonomous Driving, Heavy-Duty Vehicles}

\newcommand{\BibTeX}{\rm B\kern-.05em{\sc i\kern-.025em b}\kern-.08em\TeX}

\makeatletter
\AtBeginDocument{%
  \fancypagestyle{firstpagestyle}{%
    \fancyhf{}%
    \fancyfoot[C]{\if@ACM@printfolios\footnotesize\thepage\fi}%
  }%
}
\makeatother

\begin{document}




\maketitle 


\section{Introduction}

Autonomous driving requires real-time decision-making under uncertainty, where multiple conflicting objectives must be simultaneously balanced \cite{campbell2010autonomous,abdallaoui2023advancing}. For heavy-duty trucks, this challenge becomes even more pronounced due to their large size, high fuel consumption, long braking distances, and the severe consequences that can result from even minor control errors \cite{zhang2020safety,engstrom2018deployment}. Designing control architectures that can adaptively manage these trade-offs, prioritizing safety in dense traffic and fuel economy on open highways, and benchmarking them in controlled simulations are therefore essential for the reliable and economically viable deployment of autonomous trucks \cite{schwarting2018planning,eleonora2023potential}.

Traditional approaches to modeling autonomous driving have relied on rule-based architectures \cite{paden2016survey} and optimization-based control \cite{isaksson2016model, 7276790,en14237974}. While these methods perform well in certain tasks such as motion-planning, they often require accurate models of the environment, lack adaptability to unseen scenarios and struggle to scale with the increasing complexity of real-world traffic \cite{hu2026review}. A number of such challenges have been mitigated through Reinforcement Learning (RL) \cite{Sutton1998}, which has emerged as a promising alternative capable of learning control policies directly from interactions with traffic environments. Deep RL agents have demonstrated strong performance in tasks such as lane changing, merging, and adaptive cruise control, discovering policies that can outperform hand-crafted control strategies in complex and dynamic highway environments \cite{9351818,9304614,10386803}. However, most RL approaches in autonomous driving rely on a single scalar reward function that combines multiple objectives such as safety, comfort, and energy efficiency \cite{zhu2020safe,pathare2025tactical}. Although this simplification facilitates training, it constrains the learned policy to a fixed trade-off between objectives. 
A recent study \cite{pathare2025tactical} has investigated this paradigm in detail for heavy-duty vehicles. The authors propose a hierarchical control architecture where RL handles high-level tactical decisions using reward functions incorporating safety, efficiency, and total operational cost, while low-level control is managed by physics-based models. While this study established the practical feasibility of RL for tactical driving, it also reveals a key limitation: Agents learn stable behaviors with simpler safety-focused rewards, but struggle to optimize multiple objectives jointly, requiring careful manual weight tuning. This is not desirable in practice as it can be tedious and difficult for domain engineers to find optimal weights. Similar challenges have also been reported in other studies, where competing objectives are difficult to balance within a single reward signal and lead to poor generalization~\cite{10588385,KNOX2023103829}. 

Multi-Objective Reinforcement Learning (MORL) is designed to address precisely these challenges by representing multiple objectives as a vector-valued reward rather than combining them into a single scalar value \cite{hayes2022practical, liu2022multi}. The central task in MORL is to efficiently explore the preference space and construct a set of solutions that captures the relevant trade-offs between multiple, potentially conflicting objectives. Existing MORL algorithms differ in how they construct and represent solutions across the preference space. Preferences specify how objectives are traded off and can be expressed through a utility function; a common case is linear scalarization, where a weight vector assigns relative importance to each objective \cite{roijers2013survey, hayes2022practical}. Single-policy methods learn a policy for a specific preference, typically determined by a known utility function, whereas multi-policy methods learn multiple policies representing different trade-offs and allow preferences to be selected after training \cite{roijers2013survey, hayes2022practical, liu2022multi}. Additionally, MORL methods employ different mechanisms for identifying and improving solutions. Prediction-Guided MORL (PGMORL)~\cite{xu2020prediction}, for example, uses evolutionary search guided by predicted performance improvements to select promising preferences for training a multi-objective version of Proximal Policy Optimization (PPO) \cite{schulman2017proximal}. Gradient-Adaptive Constrained Policy Optimization (GCPO)~\cite{zhou2023gradient} adaptively adjusts policy gradients to prioritize under-optimized objectives while accounting for cost constraints. Distributional Pareto-Optimal MORL (DPMORL)~\cite{cai2023distributional}, in turn, models distributions over multi-objective returns and extends Pareto optimality to these distributions, allowing uncertainty in the returns to be considered explicitly. For the problem of constructing a set of solutions across different linear preferences, a particularly well-suited approach is Generalized Policy Improvement with Linear Support (GPI-LS)~\cite{sample_eff}, which combines targeted exploration of the preference space with the reuse of previously learned policies. The method iteratively identifies preference weights for which the current policy set can be most improved, directing training toward trade-offs that are not yet sufficiently represented. Notably, a benchmarking study providing a unified toolkit for MORL evaluation  \cite{felten2023toolkit} reported competitive performance of GPI-LS and its model-based extension, GPI-PD, in several established MORL benchmark environments compared with other state-of-the-art methods.

Despite the inherently multi-objective nature of autonomous driving, MORL has received comparatively limited attention in this domain. Early work investigated multi-objective RL for highway and urban driving, considering objectives such as safety, efficiency, and comfort~\cite{xu2018reinforcement, li2018urban, hu2021multi}. A more recent work~\cite{he2023towards} proposed an actor--critic MORL framework for preference-conditioned autonomous-driving decisions, demonstrating adaptation to different trade-offs between energy consumption and travel efficiency, while Surmann et al.~\cite{surmann2025multi} considered adaptive policies that modify driving behavior according to user preferences. These studies, however, have primarily considered passenger vehicles and do not explicitly account for the operational constraints and cost structure of heavy-duty transportation. So far, multi-objective RL has been applied to heavy-duty vehicles for longitudinal control and platoon coordination. Jia et al.~\cite{jia2025multi} proposed an MORL method for velocity control of battery-electric trucks in car-following scenarios, balancing safety, efficiency, comfort, and battery degradation, with a mixture-of-experts model used to learn the weights of the reward components. The objectives are still combined into a single scalar reward, so the result is a policy for one learned trade-off rather than a set of policies across preferences. Liu et al.~\cite{liu2026hmp} approximate a Pareto front for truck platooning, where a platoon-level agent selects acceleration and lane-change actions, with each policy trained separately for a different scalarization weight. Their learned objectives, however, are limited to speed tracking and time-to-collision-based robustness, while energy is optimized only at the planning level through platoon formation, so the learned driving policies are not energy-aware. To the best of our knowledge, existing MORL approaches for heavy-duty trucks do not learn a single preference-conditioned policy that generalizes across trade-offs, nor do they incorporate an operational cost structure which is central to commercial trucking.

In this work, we address this research gap and propose a preference-conditioned MORL framework for tactical decision making of heavy-duty trucks. The framework learns a policy that generalizes across different trade-offs among safety, energy efficiency and time efficiency, each defined in terms of practical operational costs. We build on GPI-LS \cite{sample_eff}, motivated by its competitive performance in the benchmark study \cite{felten2023toolkit} and extend it to a policy-gradient framework. We adopt Proximal Policy Optimization (PPO) \cite{schulman2017proximal}  as the underlying RL algorithm due to its strong empirical performance in tactical decision making tasks for heavy-duty vehicles ~\cite{10386803, pathare2025tactical} and its broader success in large-scale policy optimization settings. To enable multi-objective learning, we propose a multi-objective PPO (MOPPO) architecture with preference-conditioned actor and critic networks. Unlike the multi-objective PPO variant used in PGMORL~\cite{xu2020prediction}, where objective-specific representations are maintained only in the critic, we also keep objective-specific structure in the actor by learning per-objective action logits.  This preserves the structure of individual objectives in both the policy and value estimates. The framework design is intended to support generalization across preferences and the reuse of experience collected under different trade-offs, enabling stable and sample-efficient learning.

We validate the proposed framework on a realistic highway driving scenario for adaptive cruise control and lane change decision making in an autonomous truck. Using a high-fidelity microscopic traffic simulator, we demonstrate that the proposed method produces smooth and interpretable Pareto frontiers and enables preference-aware policy selection. This approach can accommodate varying operational priorities without retraining, for instance across operating contexts with different energy and labor costs. Our main contributions are as follows:
\begin{enumerate}
\item We formulate a multi-objective tactical decision-making problem for heavy-duty trucks, with safety, energy, and time objectives quantified in terms of practical operational costs.
\item We propose a preference-conditioned MORL framework that extends GPI-LS from value-based learning to a policy-gradient setting. Within this framework, we develop a multi-objective PPO (MOPPO) with objective-specific value functions and action representations.
\item We integrate the learned policy into a hierarchical decision-making architecture with kinematics-aware action masking that limits exploration to feasible actions and prevents unsafe behaviors.
\item We evaluate the framework under diverse traffic densities, including a comparison against an analytical optimal model. The results show that the trained policy generalizes across different preferences over safety, energy, and time, enabling preference-aware driving behavior without retraining.
\end{enumerate}

The proposed MORL framework and the custom RL environment for autonomous truck driving are released as open source,\footnote{Source Code: \url{https://github.com/deepthi-pathare/morl_and_sumo_gym_env}}  enabling reproducibility and facilitating future research in multi-objective decision making and autonomous truck driving.


\section{Problem Formulation}

\subsection{Tactical Decision making in highway traffic}

We study the problem of multi-objective tactical decision making for a heavy-duty truck operating in a stochastic highway environment. The objective is to learn policies for adaptive cruise control and lane changes that capture different trade-offs between safety, time efficiency, and energy efficiency.

We consider a three-lane highway segment environment populated by mixed traffic, including passenger cars and trucks. The ego vehicle is modeled as a tractor–semitrailer combination with realistic dynamics. The ego vehicle is tasked to reach a target position specified along the fixed-length highway segment within a maximum horizon. An episode terminates when the target is reached, a collision occurs, or the maximum horizon is reached.  

The environment is implemented using the open-source traffic simulator SUMO (Simulation of Urban MObility) \cite{SUMO2018}, which provides realistic microscopic vehicle dynamics. To maintain a stationary traffic distribution around the ego vehicle, a moving window is used, with vehicles dynamically re-spawned at the boundaries. Traffic density is varied across experiments to assess robustness under different congestion levels. Additional details of traffic modeling with illustrations and parameter values are provided in \autoref{app:env}.

\subsection{Multi-objective Markov Decision Process} \label{sec:momdp}

The tactical decision-making problem is inherently multi-objective: strategies that improve travel time often increase energy consumption or safety risk, while conservative driving may degrade operational efficiency. 

We formulate the problem as a Multi-Objective Markov Decision Process (MOMDP) \cite{roijers2013survey}, defined by the tuple  $\mathcal{M} = (\mathcal{S}, \mathcal{A}, p, \gamma, \mathbf{r})$
where $\mathcal{S}$ and $\mathcal{A}$ denote the state and action spaces, respectively, and 
$p(\cdot|s,a)$ is the transition probability distribution over next states given the current state--action pair $(s,a)$. 
The reward function $\mathbf{r}: \mathcal{S} \times \mathcal{A} \times \mathcal{S} \rightarrow \mathbb{R}^d$ is vector-valued, with $d$ components corresponding to distinct objectives. 
The agent’s experience thus consists of transitions $(s_t, a_t, s_{t+1}, \mathbf{r}_{t+1})$, where 
$\mathbf{r}_t = (r_t^{(1)}, \ldots, r_t^{(d)})$ quantifies the instantaneous contributions to each objective. 
$\gamma \in [0,1)$ is a discount factor.

The overall decision making architecture is hierarchical and is adopted from a recent work \cite{pathare2025tactical}; however, unlike their single-objective decision making agent, we use multi-objective learning. In this hierarchical architecture, the learned decision making agent selects high-level tactical actions, while model-based low-level controllers execute the corresponding low-level longitudinal and lateral maneuvers. At decision step $t$, the state contains observations of the ego truck and surrounding vehicles within its sensor range. The ego state includes its longitudinal position and speed, lane number and lane-change state, indicator status, vehicle dimensions, and distance to the leading vehicle. Each surrounding-vehicle state contains its relative longitudinal and lateral position, relative longitudinal speed, lane number and lane-change state, indicator status, and vehicle dimensions.

The agent selects an action from eight discrete high-level tactical actions, namely setting short ($1s$), medium  ($2s$) or long  ($3s$) desired time gap, increasing or decreasing the desired speed by $1\,\mathrm{m/s}$; maintaining the current desired speed and time gap; and initiating a lane change to the left or right. Longitudinal actions are executed by an Intelligent Driver Model (IDM)-based controller~\cite{treiber2000congested}, while lateral actions are executed using SUMO's LC2013 lane-change model~\cite{dlr89233}. The transition distribution $p$ is induced by the vehicle dynamics, low-level controllers, and stochastic behavior of surrounding traffic. Full details of the architecture and environment specification are provided in \autoref{app:arch}.

The agent optimizes three primary objectives that reflect the essential trade-off in highway driving between safety, time efficiency and energy efficiency:
\begin{enumerate}
    \item \textbf{Safety:} Avoid collisions and successfully reach the target within a finite horizon.
    \item \textbf{Time Efficiency:} Minimize the driver cost, which is a function of travel time, encouraging the agent to reach the target as quickly as possible.
    \item \textbf{Energy Efficiency:} Minimize the energy cost, encouraging the agent to adopt energy efficient driving maneuvers.
\end{enumerate}

These objectives jointly define a three-dimensional reward vector, with each objective expressed in monetary terms, given by:
\begin{align} \label{eq:rew}
    & \mathbf{r_t} = [I_{tar}R_{tar} - I_cP_c, -C_{dr}\Delta t, -C_{el}e_t]^\mathrm{T}
\end{align}
where $I$ is an indicator function, $R_{tar}$ is the reward for reaching the target, quantified using the revenue of the journey, $P_c$ is the penalty for collision, corresponding to the cost incurred during a collision, $C_{dr}$ is the driver cost per second, $\Delta t$ is the duration of a timestep, $C_{el}$ is the energy cost per $kwh$ and $e_t$ is the energy consumed in $kwh$ at time step $t$.
Detailed computations and parameter values are provided in \autoref{app:reward}.

\subsection{Multi-Objective Reinforcement Learning} \label{sec:morl}
To explicitly capture trade-offs between the objectives defined above, we learn a set of preference-conditioned policies using an MORL framework.

A policy $\pi : \mathcal{S} \to \mathcal{A}$ defines the agent’s decision rule, mapping states to actions, and value function of the policy is defined as,
\begin{align}
\mathbf{V}^\pi(s) = \mathbb{E_{\pi}} \left[ \sum_{k=0}^\infty \gamma^k \mathbf{r}_{t+k+1} \mid s_t = s \right].
\end{align}
where the value function $\mathbf{V}^\pi(s) \in \mathbb{R}^d$ is vector valued.

Optimality is defined in terms of Pareto dominance: a policy 
$\pi'$ dominates $\pi$ if it performs at least as well in all objectives and strictly better in at least one. The set of non-dominated value vectors forms the Pareto frontier, representing all achievable trade-offs beyond which improvement in one objective necessarily degrades another. MORL aims to approximate this frontier rather than identifying a single optimal policy.

User preferences are incorporated via a scalarization function or utility function, $u : \mathbb{R}^d \rightarrow \mathbb{R}$, which maps the multi-objective value vector to a scalar utility according to user-defined preferences. 
We adopt linear scalarization,
\begin{align}
u(\mathbf{V}^\pi; \mathbf{w}) = \mathbf{w}^\top \mathbf{V}^\pi = \sum_{i=1}^d w_i V_i^\pi,
\end{align}
where the weight vector $\mathbf{w}$ lies on the unit simplex. Each weight vector defines a single-objective optimization problem with scalarized rewards $\mathbf{r}_{\mathbf{w}}(s,a,s') = \mathbf{w}^\top \mathbf{r}(s,a,s')$.
The set of policies that maximize the scalarized return for some $\mathbf{w}$ forms the Convex Hull (CH), and the minimal subset containing one optimal policy per weight vector is the Convex Coverage Set (CCS), providing a compact approximation of all linearly Pareto-optimal solutions \cite{roijers2013survey}.

In our setting, the objective is to approximate CCS that optimizes across different preferences $\mathbf{w}$, capturing different trade-offs between safety, time efficiency, and energy efficiency. This motivates the need for an efficient mechanism to explore the multiobjective preference space, which we address in the following.





\section{Methodology}\label{sec:meth}
\subsection{GPI-based multi-objective reinforcement learning} \label{sec:weight_sel}


GPI-LS~\cite{sample_eff} is a value-based MORL method that combines Generalized Policy Improvement (GPI) with Linear Support (LS) to iteratively construct a coverage set across linear preferences. LS identifies preference vectors for which the current set of policies has the greatest potential for improvement, while GPI reuses the action-value functions of previously learned policies to improve policy selection and evaluate candidate preferences.

We extend this framework from its original value-based formulation to a policy-gradient setting. We retain the linear-support mechanism for selecting preference vectors, but replace value-based policy optimization with the multi-objective PPO (MOPPO) framework introduced in Section~\ref{sec:moppo}. Since standard GPI relies on action-value functions that are not directly available in PPO, we further introduce a policy-based approximation of GPI using the action distributions of the learned policies. Algorithm~\ref{alg1} summarizes the resulting procedure.

The key idea is to iteratively construct a set of policies ${\Pi} = \{\pi(a|s, \mathbf{w})\}$ conditioned on preference (weight) vectors and approximate the CCS with the associated value vectors $\boldsymbol{V}$.
At each iteration, the algorithm selects a weight vector $\mathbf{w}$ and learns a policy $\pi_{\mathbf{w}}$ that optimizes the corresponding scalarized objective. 
The resulting policy and its value vector are then added to the existing sets, refining the approximation of the CCS. 
\begin{algorithm}[h!]
\caption{GPI Linear Support (GPI-LS) with Multi-Objective PPO}
\label{alg1}
\begin{algorithmic}[1]
\Require MOMDP $M$
\State Initialize: Weight support $\mathcal{M} \gets \{\}$, Value vectors $\mathcal{V} \gets \{\}$
\State $(\pi_{\mathbf{w}}, v^{\pi_\mathbf{w}}) \gets \text{MOPPO}(\mathbf{w} = [1,0,\ldots,0]^\top)$
\State $\mathcal{V} \gets \{v^{\pi_\mathbf{w}}\},\quad \mathcal{M} \gets \{\mathbf{w}\}$
\For{$i = 1$ to $N$}
    \State $\mathcal{W}_{\text{corner}} \gets \text{CornerWeights}(\mathcal{V}) \setminus \mathcal{M}$
    \State $\mathbf{w} \gets \arg\max_{\mathbf{w} \in \mathcal{W}_{\text{corner}}} \!\left(\hat v_{\mathbf{w}}^{\mathrm{opt}} - \max_{\pi \in \Pi} v_{\mathbf{w}}^{\pi}\right)$
    \State $\mathcal{M}' \gets \text{Unique} ~\!\big(\mathcal{M} \cup \text{TopK}(\mathcal{W}_{\text{corner}}) \cup \{\mathbf{w}\}\big)$
    \State $(\pi_\mathbf{w}, v^{\pi_\mathbf{w}}, \text{done}) \gets \text{MOPPO}(\mathbf{w}, \mathcal{M}')$
    \State Add $\{{\mathbf{w}}' \in \mathcal{M}'\}$ to $\mathcal{M}$ and $\{v^{\pi_{\mathbf{w}'}} \mid {\mathbf{w}}' \in \mathcal{M}'\}$ to $\mathcal{V}$
    \State $\mathcal{V}, \mathcal{M} \gets \text{RemoveDominated}(\mathcal{V}, \mathcal{M})$
\EndFor
\end{algorithmic}
\end{algorithm}

Weight selection at each iteration is guided by the concept of corner weights~\cite{sample_eff}. These weights, denoted by $\mathcal{W}{\text{corner}} \subset \mathbb{R}^{d}$, are derived from the vertices of a polyhedron $P$ defined as
\begin{align} P = \left\{\mathbf{x} \in \mathbb{R}^{d+1} \,\middle|\, \mathbf{V}^{+} \mathbf{x} \leq \mathbf{0},\; \sum_i w_i = 1,\; w_i \geq 0,\; \forall i \right\}, \end{align}
where $\mathbf{V}^{+}$ is a matrix containing multi-objective value vectors $\mathcal{V}=\left\{\mathrm{v}^{\pi_i}\right\}_{i=1}^n$ of the $n$ trained policies $\Pi=\left\{\pi_i\right\}_{i=1}^n$, augmented with a column of $-1$s. Each vector $\mathbf{x} = (w_1, \ldots, w_d, v_{\mathbf{w}}) \in P$ represents a candidate weight vector $\mathbf{w}$ together with its associated scalarized value $v_{\mathbf{w}}$. Let $\Delta(\mathbf{w}, \Pi) = v_{\mathbf{w}}^* - \max_{\pi \in \Pi} v_{\mathbf{w}}^\pi$ denote the utility loss for a weight vector $\mathbf{w} \in \mathcal{W}$, i.e., the difference between the value of the optimal policy for $\mathbf{w}$ and the best value achievable using policies in $\Pi$. Then any maximizer $\mathbf{w} \in \arg\max_{\mathbf{w} \in \mathcal{W}} \Delta(\mathbf{w}, \Pi)$ corresponds to a corner weight of $\mathcal{V}$ \cite{roijer_phd}.

We estimate the optimal policy for computing above utility loss using Generalized Policy Improvement (GPI) \cite{gpi}. For value-based algorithms, GPI policy is defined as:
\begin{align}
\pi^{\mathrm{GPI}}(s ; \mathbf{w}) \in \arg \max _{a \in \mathcal{A}} \max _{\pi \in \Pi} q_{\mathbf{w}}^{\pi}(s, a)
\end{align}

We heuristically adapt this method to estimate the optimal policy in a PPO-based setting by maximizing scalarized action probabilities across the learned policies as given below:
\begin{align}
\hat\pi^{\mathrm{opt}}(s ; \mathbf{w}) \in \arg \max _{a \in \mathcal{A}} \max _{\pi \in \Pi} \pi(a \mid s, \mathbf{w})
\end{align}

We iteratively select the corner weight $\mathbf{w} \in \mathcal{W}_\text{corner}$ that guarantees maximum possible improvement,
\begin{align}
\mathbf{w} \gets \arg\max_{\mathbf{w} \in \mathcal{W}_{\text{corner}}} \!\left(\hat v_{\mathbf{w}}^{\mathrm{opt}} - \max_{\pi \in \Pi} v_{\mathbf{w}}^{\pi}\right)
\end{align}

 where $\hat v_{\mathbf{w}}^{\mathrm{opt}}$ is the scalarized value of estimated optimal policy. Then, the policy is updated using this $\mathbf{w}$ to maximize the scalarized return.
We choose PPO-based framework to optimize the policy $\pi_{\mathbf{w}}$ at every iteration as described in the following section.
\subsection{Multi-objective proximal policy optimization (MOPPO)} \label{sec:moppo}
\begin{figure*}[htb!]
\centering
\includegraphics[width=0.85\textwidth]{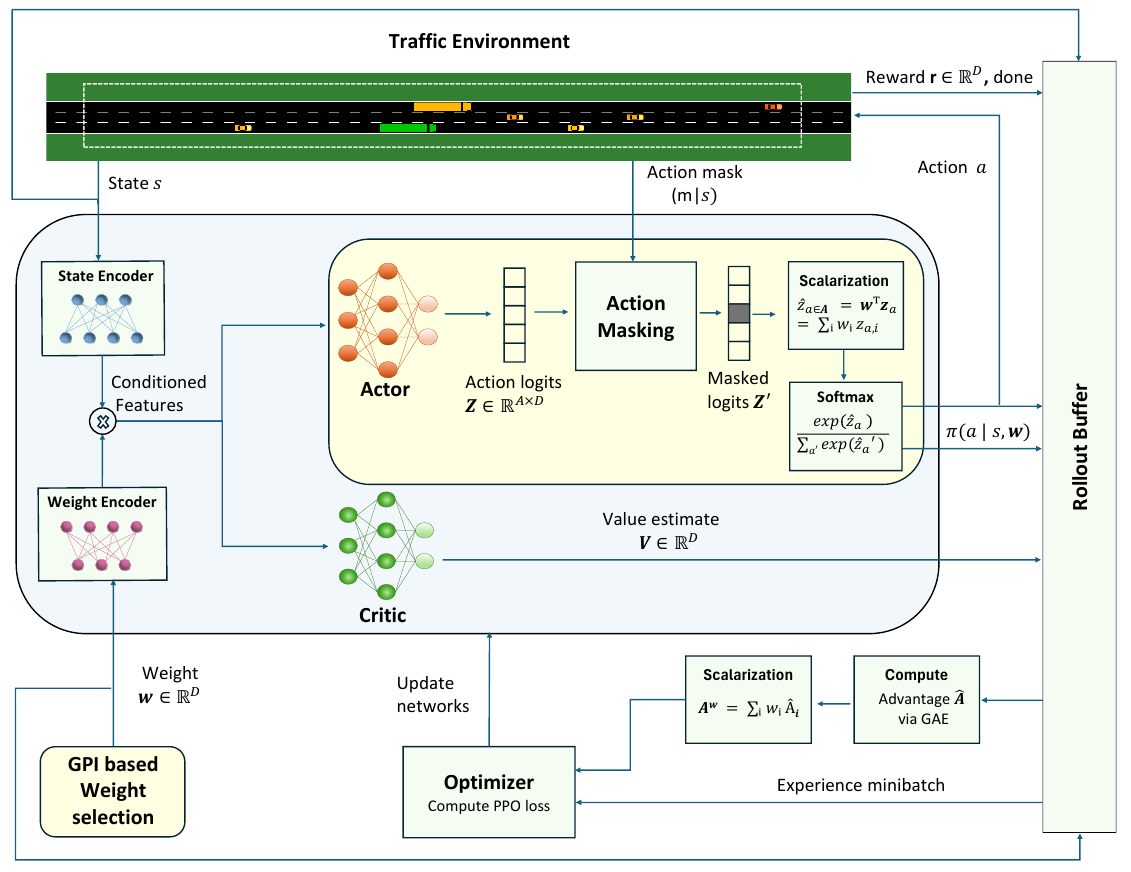}
\caption{MOPPO framework interacting with the traffic environment. The framework iteratively trains the actor and critic networks using the preference vector provided by the GPI-based weight selection algorithm described in Section \ref{sec:weight_sel}. 
}
\Description{MOPPO framework interacting with the traffic environment. The framework iteratively trains the actor and critic networks using the preference vector provided by the GPI-based weight selection algorithm described in Section \ref{sec:weight_sel}.}
\label{fig:moppo}
\end{figure*}

In Algorithm \ref{alg2}, we extend the PPO \cite{schulman2017proximal} to handle multi-objective learning by incorporating weight-conditioned actor and critic networks. Our algorithm follows the standard clipped PPO update, but modifies the policy and value function estimation to support multi-dimensional rewards and scalarization using weight vectors. See  \autoref{fig:moppo} for a schematic overview of the proposed framework.

\begin{algorithm}[h!]
\caption{MOPPO Training (single iteration)}
\label{alg2}
\begin{algorithmic}[1]
\renewcommand{\algorithmicrequire}{\textbf{Require:}}
\Require policy parameters $\boldsymbol{\theta}$, value parameters $\boldsymbol{\phi}$, steps per iteration $N_S$, selected corner weight $\mathbf{w}$ and weight support $\mathcal{M}$ (from Algorithm \ref{alg2})

\State \textbf{Initialize:} replay buffer $\mathcal{D} \leftarrow \emptyset$, $\mathbf{w}_t \gets \mathbf{w}$

\For{each environment step $t = 1, \ldots, N_S$}
    \State \parbox[t]{\dimexpr\linewidth-\algorithmicindent}{Sample action $a_t \sim \pi_\theta(a \mid s_t, \mathbf{w}_t)$ and estimate value $\mathbf{v}_t = V_\phi(s_t, \mathbf{w}_t)$}
    
    \State \parbox[t]{\dimexpr\linewidth-\algorithmicindent}{Execute $a_t$ in the environment to obtain $(\mathbf{r}_{t+1}, s_{t+1}, {done}_{t+1})$}
    
    \State \parbox[t]{\dimexpr\linewidth-\algorithmicindent}{Store
    $(s_t, a_t, \log \pi_\theta(a_t \mid s_t, \mathbf{w}_t), \mathbf{r}_{t+1}, {done}_{t+1}, \mathbf{v}_t, \\
    \mathbf{w}_t)$ in $\mathcal{D}$}
    
    \If{episode terminates}
        \State Sample new $\mathbf{w}_t \sim \mathcal{M}$ and reset environment
    \EndIf
\EndFor

\State Compute vector advantage estimates $\hat{\mathbf{A}}_t$ using GAE$(\gamma, \lambda)$
\State Scalarize advantages: $A_t^{(s)} = \mathbf{w}_t^\top \hat{\mathbf{A}}_t = \sum_{i=1}^{d} w_{t,i}\,\hat{A}_{t,i}$

\For{each update epoch}
    \State \parbox[t]{\dimexpr\linewidth-\algorithmicindent}{Sample minibatches from $\mathcal{D}$ and update $(\theta, \phi)$ by maximizing: 
    \[ \mathcal{L}_{t}^{\textit{CLIP+VF+S}}(\boldsymbol{\theta}, \boldsymbol{\phi}) = \mathcal{L}_{t}^{\textit{CLIP}}(\boldsymbol{\theta}) - c_1 \mathcal{L}_{t}^{\textit{VF}}(\boldsymbol{\phi}) + c_2 \mathcal{S}(\pi_{\boldsymbol{\theta}}) \]}
\EndFor

\State \textbf{Output:} Updated policy $\pi_{\boldsymbol{\theta}}$ and value function $V_{\boldsymbol{\phi}}$
\end{algorithmic}
\end{algorithm}

The framework consists of the following:
\begin{enumerate} [leftmargin=*]
    \item \textbf{Weight-conditioned feature extraction:} The observation features and the preference vector $\mathbf{w} \in \mathbb{R}^d$ are independently encoded using multi-layer perceptrons (MLPs), and their resulting feature representations are combined element-wise. This conditioning mechanism modulates the state representation according to the preference vector, activating different feature subspaces and allowing the policy network to adapt its behavior to the specified trade-off weights.
    \item \textbf{Multi-objective actor:} The actor outputs a set of action logits $\mathbf{Z}(a \,|\, s) \in \mathbb{R}^{|\mathcal{A}| \times d}$, one per reward dimension. These logits are scalarized by $\mathbf{w}$ to obtain a single distribution over actions, from which actions are sampled. 
    \item \textbf{Action masking:} The policy employs an action-masking mechanism conditioned on the current state to limit exploration to feasible actions and prevent unsafe decisions. This approach helps stabilize training and improves sample efficiency. The environment provides a binary mask identifying feasible actions, and logits corresponding to infeasible actions are assigned a large negative value prior to the softmax operation, ensuring their selection probability becomes effectively zero.
    
    \item \textbf{Multi-objective critic:} The critic outputs a vector-valued estimate $V(s) \in \mathbb{R}^d$, predicting the expected return for each reward dimension. 
    \item \textbf{Rollout buffer:} 
    At each iteration, MOPPO trains the network based on the selected corner weight $\mathbf{w}_t$ that gives maximum improvement as mentioned in Section \ref{sec:weight_sel}. The algorithm interacts with the environment to gather trajectories conditioned on a weight, either the selected corner weight $\mathbf{w}_t$ or a weight vector sampled uniformly from weight support $\mathcal{M}$, each with probability 0.5. While the selected corner weight targets the most promising direction for improvement, sampling from $\mathcal{M}$, which contains previously selected and high-performing corner weights retains information about diverse trade-offs, facilitating exploration and generalization across the preference space.
    Each transition $(s_t, a_t, \log \pi(a_t|s_t, \mathbf{w}_t), \mathbf{r}_{t+1}, \text{done}_{t+1}, \mathbf{V}(s_t), \mathbf{w}_t)$ is stored in a rollout buffer. The collected data are then used to compute the objective function and update the networks as described next.

\item \textbf{Optimization:} The trainable parameters of the policy and value networks, denoted by $\boldsymbol{\theta}$ and $\boldsymbol{\phi}$, are updated by maximizing the PPO-style objective:
\begin{align}
\mathcal{L}_{t}^{\textit{CLIP+VF+S}}(\boldsymbol{\theta},\boldsymbol{\phi}) = \mathcal{L}_{t}^{\textit{CLIP}}(\boldsymbol{\theta}) - c_1 \mathcal{L}_{t}^{\textit{VF}}(\boldsymbol{\phi}) +
c_2 \mathcal{S}(\pi_{\boldsymbol{\theta}})
\end{align}
where $\mathcal{L}_{t}^{\text{CLIP}}(\boldsymbol{\theta})$ is the clipped surrogate objective, $\mathcal{L}_{t}^{\text{VF}}(\boldsymbol{\phi})$ is the squared-error value loss, $\mathcal{S}(\pi_{\boldsymbol{\theta}})$ denotes entropy bonus, and $c_1, c_2$ are weighting coefficients for the value loss and entropy bonus, respectively.

\begin{equation}
\begin{aligned}
\mathcal{L}_{t}^{\textit{CLIP}}(\boldsymbol{\theta})
&= \mathbb{E}\Bigg[
\min\Bigg(
\rho_t(\boldsymbol{\theta}) A_t^{(s)}, \\
&\qquad\operatorname{clip}\big(
\rho_t(\boldsymbol{\theta}),
1-\epsilon,
1+\epsilon
\big) A_t^{(s)}
\Bigg)
\Bigg]
\end{aligned}
\tag{9}
\end{equation}

where
\begin{align}
\rho_t(\boldsymbol{\theta}) =
\frac{\pi_{\boldsymbol{\theta}}(a_t \mid s_t, \mathbf{w}_t)}
{\pi_{{\boldsymbol{\theta}}_{\text{old}}}(a_t \mid s_t, \mathbf{w}_t)}, 
\quad
A_t^{(s)} = \mathbf{w}_t^\top \hat{\mathbf{A}}_t.
\end{align}
$\hat{\mathbf{A}}_t \in \mathbb{R}^d$ denotes the multi-objective advantage computed using Generalized Advantage Estimation (GAE-$\lambda$) \cite{schulman2015highdimensional}.
\end{enumerate}

\subsection{Safety filter for lane changes via action masking} \label{meth:safetyfil}
To complement the safety objective in the multi-objective reward, we impose a rule-based safety filter that restricts the RL agent from executing lane change actions that violate basic kinematic feasibility conditions. Thus, safety is considered at two complementary levels: the reward function encourages the policy to optimize safety jointly with the other objectives, while the safety filter acts as a hard constraint on the set of executable lane change actions. Any lane change action that violates these constraints is masked out as described in Section \ref{sec:moppo}, restricting the policy's exploration to feasible actions.

The filtering considers both road-boundary constraints and gap-based safety constraints. A lane change to the left when the ego vehicle is on the left-most lane, and a lane change to the right when the vehicle is on the right-most lane, are filtered out. Candidate lane changes are further evaluated using a kinematic, gap-based safety filter that explicitly accounts for finite vehicle dimensions, relative velocities, and the fact that the ego vehicle occupies both the current and target lanes during the maneuver. A lane change is permitted only if the available gaps with respect to the ego vehicle are sufficient in all relevant interactions, namely: (i) the gap to the leading vehicle in the current lane during lane exit, (ii) the gap to the leading vehicle in the target lane at time of lane entry and over the lane change duration, and (iii) the gap to the following vehicle in the target lane at lane entry. This safety filter enforces feasible merging conditions, thereby reducing the risk of collisions. The detailed formulation of the safety constraints is provided in \autoref{app:safety}.

\section{Experimental results} \label{sec:res}

In this section, we present results from experiments conducted using the proposed MORL framework in the multi-objective RL environment that we developed for highway truck driving. Our GPI-LS MOPPO algorithm is implemented on top of the MORL-Baselines toolkit \cite{felten2023toolkit}. 
All experimental settings including hyperparameter values are given in \autoref{app:exp_sett}.

\subsection{Pareto-fronts in different traffic settings} \label{sec:res_pareto}

\begin{figure*}[h]
\centering
\includegraphics[width=0.98\textwidth]{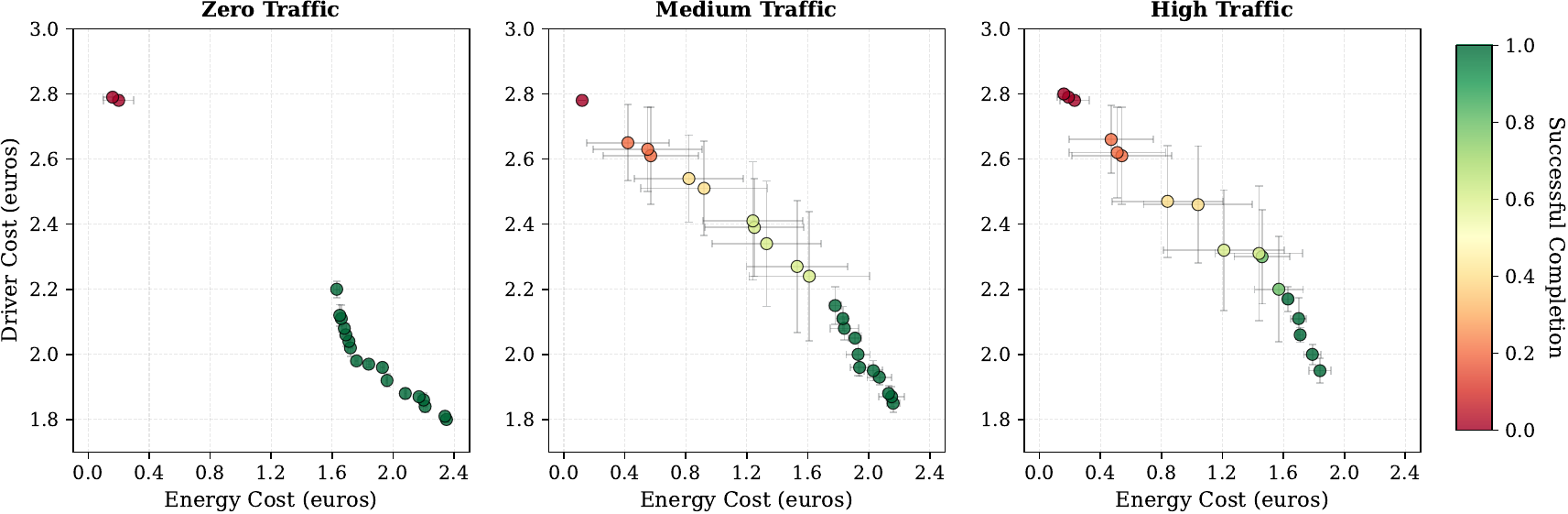}
\caption{{Pareto front produced by the GPI-LS MOPPO framework, showing the trade-off between driver cost and energy cost across three traffic settings, along with the success rate. Markers represent mean driver and energy costs over five evaluation episodes, with error bars indicating the standard error.}} 
\Description{Pareto front produced by the GPI-LS MOPPO framework, showing the trade-off between driver cost and energy cost across three traffic settings, along with the success rate. Markers represent mean driver and energy costs over five evaluation episodes, with error bars indicating the standard error.}
\label{fig:pareto}
\end{figure*}

We begin by examining the trade-offs learned by the agent trained with GPI-LS MOPPO algorithm under varying traffic densities. \autoref{fig:pareto} shows the Pareto-optimal policies for zero, medium, and high traffic scenarios, illustrating the trade-off between energy cost and driver cost together with their corresponding successful completion rates. Each Pareto front is obtained by evaluating the trained agent over 500 equally spaced weight vectors, with each policy evaluated over 5 episodes. The markers represent the mean driver and energy costs of the non-dominated policies, while the error bars indicate the standard error of the mean (SEM) across the five evaluation episodes. We include only feasible policies with $0\%$ collisions in the Pareto fronts, since early collisions may result in very low cost values, thereby distorting the Pareto structure. 

For medium traffic density, we use 0.015 vehicles per meter, which corresponds to 7 vehicles (6 cars and 1 truck) based on the chosen moving window size and heterogeneous vehicle ratio reported in \autoref{app:env}. For high traffic density, we use 0.03 vehicles per meter, corresponding to 13 vehicles (11 cars and 2 trucks). For all three settings, the GPI-LS MOPPO framework is trained for 100 iterations (i.e., $N = 100$ in Algorithm~\ref{alg1}), with each iteration consisting of 10{,}000 training steps (i.e., $N_S = 10{,}000$ in Algorithm~\ref{alg2}), leading to a total timesteps of $1 \times 10^6$. Across all traffic settings, a clear and interpretable Pareto structure emerges, demonstrating that the learned policies successfully capture the fundamental conflict between energy cost and driver cost. 


In the zero-traffic scenario, the deterministic environment makes reachability largely dependent on the policy under evaluation, so episodes typically either succeed or fail for a fixed policy. The absence of non-dominated policies in the intermediate region of the Pareto front follows from this structure. Policies with low cruising speeds cannot reach the target within the maximum RL steps, leading to an almost constant driver cost due to episode time-out (with only minor variation from lane changes). As a result, low-cost policies cluster in the cost space, effectively preserving only those with minimal energy consumption and eliminating trade-offs in the intermediate region of the Pareto front.

In contrast, when traffic is present, interactions with other vehicles impose state-dependent speed constraints that introduce variability in both travel time and energy usage. Even policies with similar average speeds can experience different patterns of acceleration, deceleration, and lane changes, resulting in a wider range of achievable cost combinations and target reachability. This leads to multiple non-dominated policies spanning a broader region of the Pareto front. The success rate remains high for a wide range of trade-offs in medium and high traffic, indicating that the proposed framework effectively learns policies that balance driver and energy costs without compromising feasibility of the task.

The error bars in the three settings also reveal a consistent pattern across the evaluated policies. Policies with either $0\%$ or $100\%$ success rate generally exhibit smaller errors, whereas policies with intermediate success rates tend to show greater uncertainty in their mean cost estimates. This is expected because intermediate success rates indicate a mixture of successful and unsuccessful episodes, which can produce greater variation in the resulting cost values.

\subsection{Comparison with analytical solution} \label{sec:res_analy}
\begin{figure*}[htb!]
\centering
\includegraphics[width=0.98\textwidth]{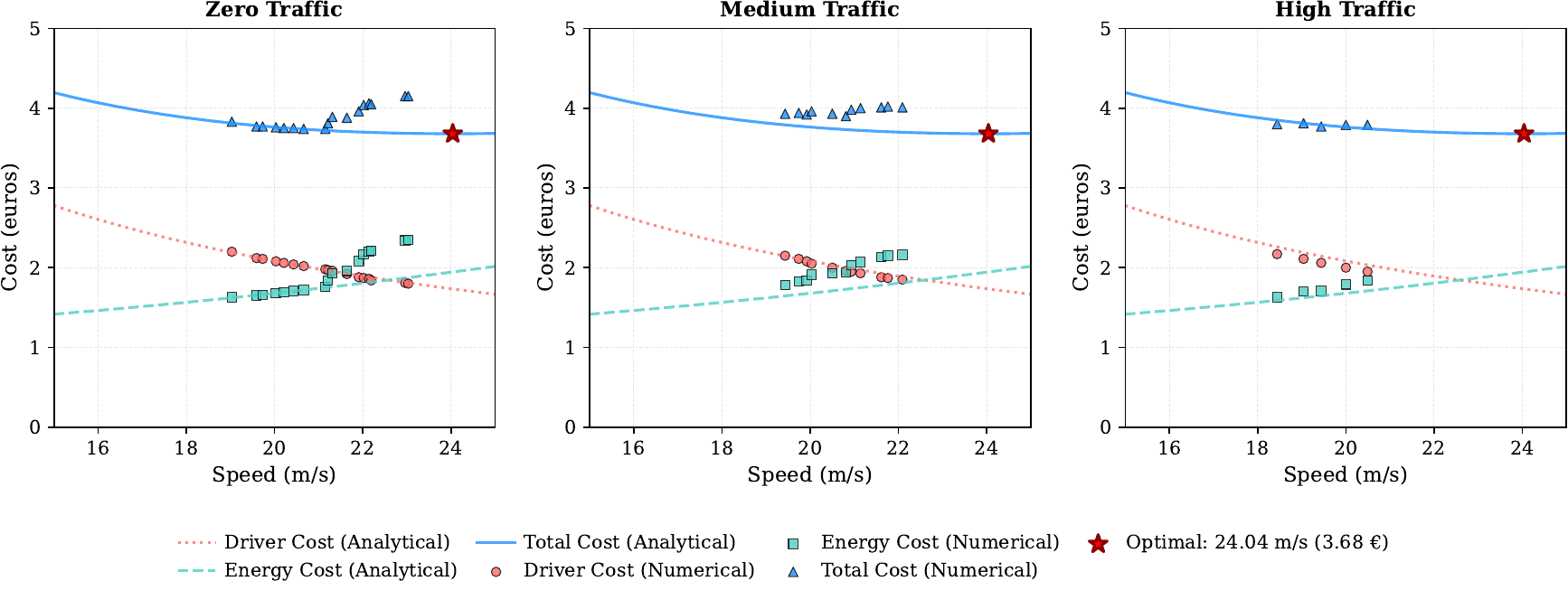}
\caption{Comparison of average speed and cost values obtained for Pareto-optimal policies by GPI-LS MOPPO, with $100\%$ success rate, against analytical values. The analytical curves show cost as a function of speed under a constant-speed assumption.} 
\Description{Comparison of average speed and cost values (driver cosr, energy cost and total cost) obtained for Pareto-optimal policies by GPI-LS MOPPO, with $100\%$ success rate, against analytical values. The analytical curves show cost as a function of speed under a constant-speed assumption.}
\label{fig:speed_cost}
\end{figure*}

\autoref{fig:speed_cost} compares the average speed and corresponding mean cost values of Pareto-optimal policies in \autoref{fig:pareto} that achieved a $100\%$ successful completion rate, against the analytical optimal speed and cost obtained under the constant-speed assumption. A detailed graph of the analytical solution, illustrating how different cost components vary as a function of speed, is provided in \autoref{fig:anal_res} in \autoref{app:results}. The analysis identifies an optimal constant speed of 24.04 m/s, corresponding to a minimum total cost of 3.68 euros for a 3000-meter trip (0.0012 euros/m) for the considered environment settings. In the zero-traffic scenario, the policies learned by the algorithm closely match this analytical baseline. The absence of traffic interactions enables the ego vehicle to maintain an approximately constant speed with minimal fluctuations throughout the episode. As a result, the average speed computed over time closely reflects the instantaneous speed, and the resulting driver cost and energy cost align well with the analytical cost model. The larger deviations in energy cost in the higher-speed regime may be partly attributed to the cubic dependence of energy consumption on speed (see \autoref{app:reward}), which amplifies small velocity variations in the resulting energy estimates.

Under medium and high traffic densities, the relationship between average speed and cost deviates increasingly from the analytical curves. In these settings, the reported average speed represents a temporal mean over highly variable speed profiles that include frequent acceleration, deceleration, and lane change maneuvers induced by interactions with surrounding vehicles. While different policies may exhibit similar average speeds, their underlying speed trajectories can differ substantially, leading to different travel times and energy expenditures. Therefore, when surrounding vehicles are added, these variations become more pronounced, resulting in small deviations between the analytically predicted costs and the observed costs from the experiments.  Moreover, as traffic density increases, there is a scarcity of policies whose average speed approaches the analytical optimum. Higher traffic density constrains the ego vehicle’s ability to accelerate and sustain high cruising speeds, limiting its ability to operate near the analytically optimal speed. Nonetheless, when costs are normalized per meter, the best achieved \textit{total cost per meter} (0.0012 euros) matches the analytical value in all three traffic settings. Detailed numerical results for all traffic conditions and policies are provided in Appendix~\ref{app:results}.
\subsection{Baseline comparison and ablation study} \label{sec:res_base}

\begin{figure*}[htb]
\centering
\includegraphics[width=0.98\textwidth]{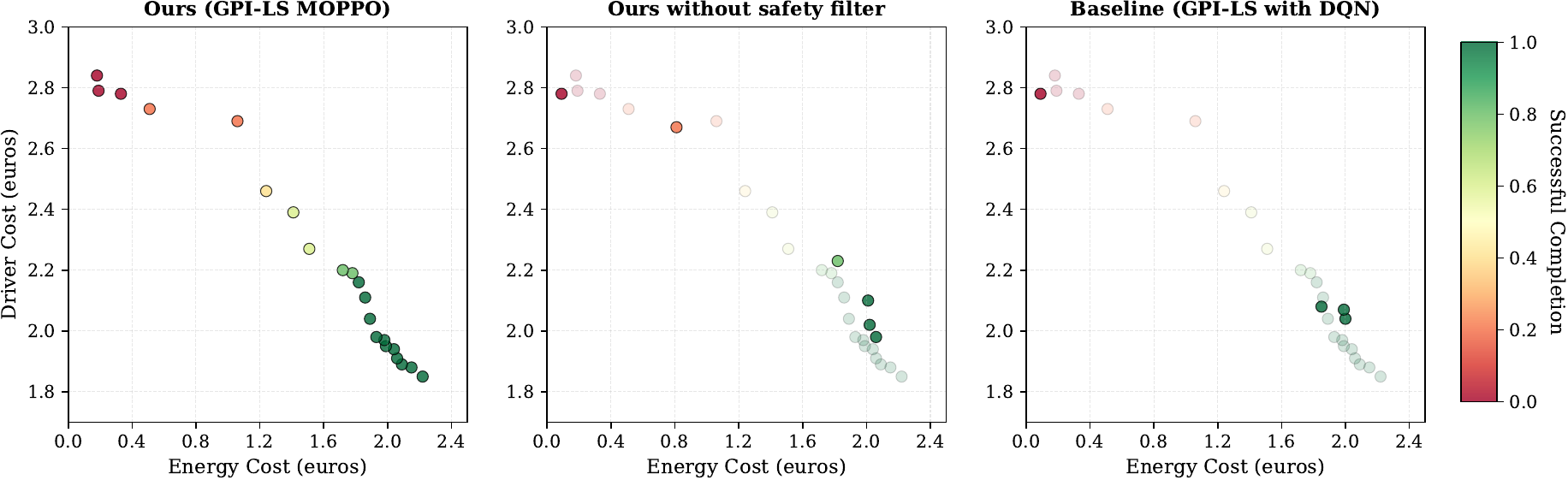}
\caption{Comparison of Pareto-frontiers produced by our method (GPI-LS MOPPO), its ablated variant without safety filter and a basline method (GPI-LS with DQN). The faded points in the last two plots correspond to the Pareto policies of our method from the first plot, added for direct comparison.}
\Description{Comparison of Pareto-frontiers produced by our method (GPI-LS MOPPO), its ablated variant without safety filter and a basline method (GPI-LS with DQN). The faded points in last two plots correspond to the policies in the first plot, added for direct comparison.}
\label{fig:pareto_comparison}
\end{figure*}

\autoref{fig:pareto_comparison} compares the Pareto fronts generated by GPI-LS MOPPO against a baseline GPI-LS with Deep Q-Network (DQN) \cite{sample_eff}. Additionally, we perform an ablation study of our method without the safety filter. The experiments are conducted in a medium traffic setting, where all three methods are trained for 50 iterations of GPI-LS with a total of $5 \times 10^5$ timesteps.  

Compared to the baseline, our method identifies a larger set of feasible policies (driving without collisions), while covering a broader range of trade-offs between driver cost and energy cost. The ablation study highlights the importance of the safety filter. Removing it reduces the number of feasible policies on the Pareto frontier. Moreover, among policies with higher success rates, our method achieves lower cost values compared to both the baseline and the ablated variant. The cost per meter is also lower than that obtained with scalarized single-objective RL in a similar setting \cite{pathare2025tactical}, which required carefully tuned weights to perform well. 

Further numerical evaluations using metrics such as \textit{expected utility}, \textit{mean utility}, and \textit{computational efficiency} also demonstrate that our method outperforms the baseline, as shown in \autoref{app:results}.

\section{Conclusion} \label{sec:conclusion}
We developed a multi-objective reinforcement learning framework for tactical decision making in heavy-duty vehicles by extending Generalized Policy Improvement with Linear Support (GPI-LS) to a policy-gradient setting using Multi-Objective Proximal Policy Optimization (MOPPO). The proposed framework combines targeted exploration of the preference space with preference-conditioned policy learning, allowing a set of driving policies to be learned across different trade-offs between safety, time efficiency, and energy efficiency. Experiments in a stochastic highway environment show that the proposed approach learns a diverse set of policies spanning different trade-offs between success rate, driver cost, and energy cost. Evaluation across different traffic densities indicates that these trade-offs persist under varying traffic conditions, while close alignment with the analytical baseline provides an additional evidence for the quality of learned solutions.

The proposed framework provides a step toward preference-aware tactical decision making for heavy-duty vehicles, where operating strategies can be selected according to application-specific priorities. The present evaluations, however, are limited to highway scenarios with predefined traffic conditions, and do not investigate how the learned policies generalize to more diverse driving environments, such as urban traffic or complex intersections. Future work will investigate more complex traffic scenarios and road geometries and richer preference models.






\clearpage
\newpage
\bibliographystyle{ACM-Reference-Format} 
\bibliography{sample}


\clearpage
\newpage
\appendix
\section{Traffic environment modeling} \label{app:env}
\begin{figure*}[h!]
\centering
\includegraphics[width=0.7\textwidth]{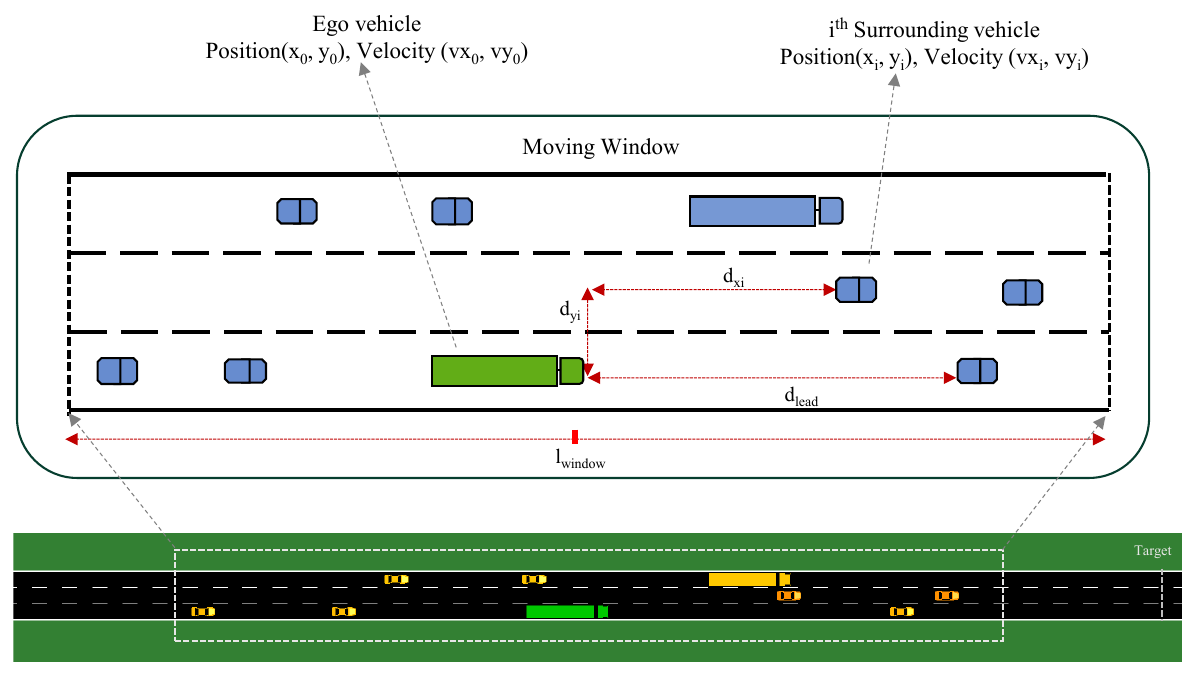}
\caption{Simulated traffic environment in SUMO with the illustration of moving window.}
\Description{Simulated traffic environment in SUMO with the illustration of moving window.}
\label{fig:env}
\end{figure*}

We created a custom RL environment for autonomous trucks in highway traffic as illustrated in \autoref{fig:env}. The ego truck highlighted in green color is controlled by our learned decision making policy whereas surrounding vehicles are simulated using the Krauss car following model \cite{Krauss1997} and LC2013 lane change model \cite{dlr89233} in SUMO. Each surrounding vehicle is assigned a maximum speed sampled from a distribution. The traffic parameters used in this study are provided in \autoref{tab:sim_params}. 

\begin{table}[h!]
\caption{Traffic simulation parameters.}
\label{tab:sim_params}
\centering
\begin{tabular}{p{5.4cm} p{2.2cm}}
\toprule
\textbf{Parameter} & \textbf{Value} \\
\midrule
Length of highway segment (l\textsubscript{road}) & 3000 m \\ 
\hline
Moving window size (l\textsubscript{window})& 400 m \\ 
\hline
Heterogeneous vehicle ratio (trucks) & 0.2 \\ 
\hline
Heterogeneous vehicle ratio (cars) & 0.8 \\ 
\hline
Car max speed distribution & $\mathcal{N}(23, 3.8)$ m/s \\ 
\hline
Truck max speed distribution & $\mathcal{N}(20, 0.8)$ m/s \\ 
\hline
Maximum speed of ego truck & 25 m/s \\ 
\hline
Maximum acceleration of ego truck & 0.1 $m/s^2$ \\ 
\hline
Maximum deceleration of ego truck & 6 $m/s^2$ \\ 
\hline
Maximum episode length & 200 \\ 
\bottomrule
\end{tabular}
\end{table}

\section{Detailed architecture of decision making framework} \label{app:arch}
\begin{figure}[h!]
\centering
\includegraphics[width=0.45\textwidth]{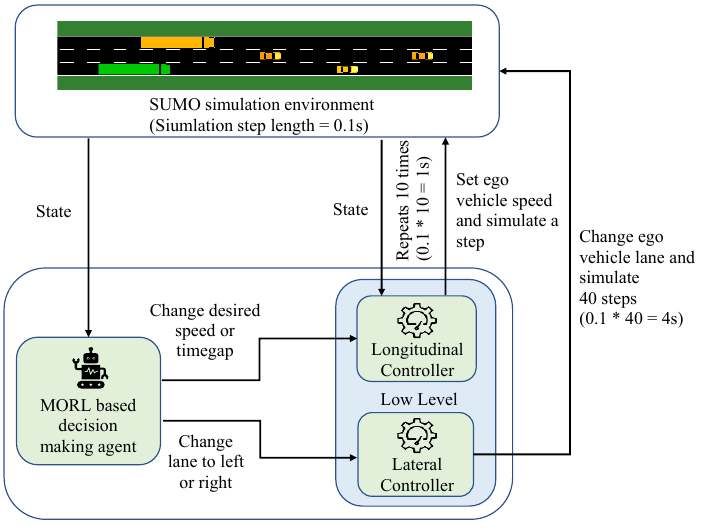}
\caption{Overview of the architecture adapted from \protect\cite{pathare2025tactical}, extended for multi-objective learning}
\Description{Overview of the architecture adapted from \protect\cite{pathare2025tactical}, extended for multi-objective learning}
\label{fig:arch}
\end{figure}

The decision making and control architecture is hierarchical integrating RL with low-level controllers as depicted in \autoref{fig:arch}. It is adapted from the paper \cite{pathare2025tactical}, extended for MORL. 

We separate the high-level and low-level decision making between MORL agent and low-level controllers. The agent has a discrete action space which includes high level decisions about longitudinal and lateral controls. The longitudinal actions include changing the desired speed or desired timegap with the vehicle. Lateral actions include changing the lane to left or right. The action space includes:
\begin{enumerate}
    \item Set short desired time gap with leading vehicle (1s)
    \item Set medium desired time gap with leading vehicle (2s)
    \item Set long desired time gap with leading vehicle (3s)
    \item Increase the desired speed by 1 m/s
    \item Decrease the desired speed by 1 m/s
    \item Maintain current desired speed and time gap
    \item Change lane to left
    \item Change lane to right
\end{enumerate}

When one of the longitudinal action is chosen it triggers the longitudinal controller which compute the acceleration/deceleration using the set desired speed and timegap. We use  Intelligent Driver Model (IDM) \cite{treiber2000congested} given by,
\begin{equation}
\begin{aligned}
& \dot{v}_\alpha=\frac{\mathrm{d} v_\alpha}{\mathrm{d} t}=a\left(1-\left(\frac{v_\alpha}{v_0}\right)^\delta-\left(\frac{s^*\left(v_\alpha, \Delta v_\alpha\right)}{s_\alpha}\right)^2\right), \\
& s^*\left(v_\alpha, \Delta v_\alpha\right)=s_0+v_\alpha T+\frac{v_\alpha \Delta v_\alpha}{2 \sqrt{a b}}
\label{eq:idm}
\end{aligned}
\end{equation}
\noindent
where $\alpha$ is the ego vehicle and $\alpha - 1$ is the leading vehicle. $v$ denotes the velocity and $l$ denotes the length of the vehicle. $s_\alpha:=x_{\alpha-1}-x_\alpha-l_{\alpha-1}$ is the net distance and $\Delta v_\alpha:=v_\alpha-v_{\alpha-1}$ is the velocity difference. $v_0$ (desired velocity), $s_0$ (minimum spacing), $T$ (desired time gap), $a$ (maximum acceleration), and $b$ (comfortable braking deceleration) are model parameters.  

A new acceleration is computed for the truck and the resulting speed is set to the vehicle every 0.1s. This process continues for a total duration of 1s, after which the RL agent chooses the next high level action. If a lateral action is chosen by the agent, the lateral controller initiates the lane change. Lane change is performed using the default LC2013 lane change model \cite{dlr89233} in SUMO. The lane width is set to $3.2m$ and the lateral speed of the truck is set to $0.8 m/s$. Hence, in total, it takes $4s$ to complete a lane change, following which RL chooses the next high level action.

The state space of RL includes the observations for ego truck and surrounding vehicles.
Following are the observations for the ego vehicle:
\begin{enumerate}
    \item Longitudinal position
    \item Longitudinal speed 
    \item Lane number
    \item Lane change state 
    \item State of left indicator 
    \item State of right indicator
    \item Length of the vehicle
    \item Width of the vehicle
    \item Target (leading) vehicle distance
\end{enumerate}

Following are the observations for each vehicle in the sensor range of the ego vehicle:
\begin{enumerate}
    \item Relative longitudinal distance from ego vehicle
    \item Relative lateral distance from ego vehicle
    \item Relative longitudinal speed with ego vehicle
    \item Lane number
    \item Lane change state
    \item State of left indicator
    \item State of right indicator
    \item Length of the vehicle
    \item Width of the vehicle
\end{enumerate}

\section{Reward computation} \label{app:reward}
As mentioned in Section \ref{sec:meth}, the reward vector consists of the following components.
\begin{align} \label{eq:rew}
    & r_t = [I_{tar}R_{tar} - I_cP_c, -C_{dr}\Delta t, -C_{el}e_t]^T
\end{align}
Here, the reward for reaching the target $R_{tar}$ corresponds to the revenue of the journey, which is computed as analytical optimal cost (see Appendix \ref{app:anal_sol}) plus a $20\%$ profit margin. The cost incurred during a collision $P_c$ is considered to be the own risk payment which is a portion of insurance claim made by the vehicle owner or the insured. $C_{el}$ is the electricity cost, $e_t$ is the electricity consumed at time step $t$, $C_{dr}$ is the driver cost and $\Delta t$ is the duration of a time step. $\Delta t$ would be $1s$ for a longitudinal action and $4s$ for a lateral action. The electricity consumed during the time step $t$ ($e_t$) is calculated as,

\begin{equation}
\begin{aligned}
\label{eq:energy}
    e_t = f_t\:v_t\:\Delta{t},
\end{aligned}
\end{equation}
where $f_t$, force at time step $t$ is given by, 
\begin{equation}
\begin{aligned}
f_t = m\:a_t + \frac{1}{2} C_d\:A_f \: \rho_{\text{air}} \: v^2 + m \: g \: C_r \\ +m \: g \: \sin (\arctan ( \frac{\text{slope}}{100}))
\end{aligned}
\end{equation}
Here $m$ is the mass of the vehicle, $C_d$ is the coefficient of air drag, $A_f$ is the frontal area, $\rho_{\text{air}}$ is the air density, $C_r$ is the coefficient of rolling resistance, $g$ is the acceleration due to gravity and $a$ is the acceleration of the vehicle at time step $t$. We use a road segment with 0 slope in this study. The parameter values are given in \autoref{tab:rew_params}.

\begin{table}[t]
\caption{Parameter values used for reward computation.}
\label{tab:rew_params}
\centering
\begin{tabular}{ll}
\toprule
\textit{Parameter} & \textit{Value} \\ 
\midrule
$R_{tar}$ & 4.41 \\
$P_c$ & 1000 \\
$C_{el}$ & 0.5 euro per kWh \\
$C_{dr}$ & 50 euro per hour \\
$m$ & 44000 kg \\
$C_d$ & 0.6 \\
$A_f$ & 10 m$^2$ \\
$\rho_{\text{air}}$ & 1.2 kg/m$^3$ \\
$g$ & 9.81 m/s$^2$ \\
$C_r$ & 0.006 \\
\bottomrule
\end{tabular}
\end{table}

\section{Safety constraints for lane change actions}\label{app:safety}
This section provides the details of safety constraints for lane change actions.

The total duration for a lane change is given by, 
\begin{align}
T_{\text{lc}} = \frac{w_{\text{lane}}}{v_{\text{lat}}},
\end{align}
where $w_{\text{lane}}$ is the lane width and $v_{\text{lat}}$ is the lateral speed of the ego vehicle.
The ego vehicle enters the target lane at,
\begin{align}
t_{\text{enter}} = \frac{w_{\text{lane}} - w_{\text{ego}}}{2 v_{\text{lat}}},
\end{align}
and fully exits the current lane at,
\begin{align}
t_{\text{exit}} = \frac{w_{\text{lane}} + w_{\text{ego}}}{2 v_{\text{lat}}}.
\end{align}
 where $w_{\text{ego}}$ is the ego vehicle's width.

The minimum required gap to a leading vehicle is given by,
\begin{align}
s_{\min}(v, \Delta v)
= s_0 + 
T_{\text{safe}} v + \frac{v \Delta v}{2 \sqrt{a_{\max} b_{\text{safe}}}},
\end{align}

as defined in IDM \cite{treiber2000congested}. Here, 
a minimum standstill gap $s_0$ and desired time headway $T_{\text{safe}}$
are used to define safety margins. $v$ is the follower speed, $\Delta v$ is relative speed with the leading vehicle, $a_{\max}$ and $b_{\text{safe}}$ denote maximum acceleration and comfortable deceleration respectively.

From the observation of environment, we identify the nearest leading and following vehicles in the current and target lanes. All longitudinal distances are measured between front bumpers. A lane change is permitted only if all of the following conditions hold:

\begin{enumerate}
\item {Front gap in current lane:}
Let $s_{\text{front,cur}}$ be the distance to the leading vehicle in the current lane. The gap must remain safe until the ego fully exits the current lane:
\begin{align}
   s_{\text{front,cur}} - (v_{\text{ego}} - v_{\text{front,cur}})\, t_{\text{exit}}
    \ge s_{\min}(v_{\text{ego}}, v_{\text{ego}} - v_{\text{front,cur}}) 
\end{align}

    \item {Front gap in target lane:}
    Let $s_{\text{front,tar}}$ denote the distance to the
    leading vehicle in the target lane.
    The gap must be safe both when the ego enters the target lane and
    at the end of the maneuver:
    \begin{align}
    s_{\text{front,tar}} - (v_{\text{ego}} - v_{\text{front,tar}})\, t_{\text{enter}}
    \ge s_{\min}(v_{\text{ego}}, v_{\text{ego}} - v_{\text{front,tar}}),
    \end{align}
    and
    \begin{align}
    s_{\text{front,tar}} - (v_{\text{ego}} - v_{\text{front,tar}})\, T_{\text{lc}}
    \ge s_{\min}(v_{\text{ego}}, v_{\text{ego}} - v_{\text{front,tar}}).
    \end{align}

    \item {Rear gap in target lane:}
    Let $s_{\text{rear,target}}$ be the distance from the ego
    to the following vehicle in the target lane.
    The gap at lane entry must satisfy
    \begin{align}
    s_{\text{rear,tar}} - (v_{\text{rear,tar}} - v_{\text{ego}})\, t_{\text{enter}}
    \ge s_{\min}(v_{\text{rear,tar}}, v_{\text{rear}} - v_{\text{ego}}).
    \end{align}

\end{enumerate}

\section{Experimental settings} \label{app:exp_sett}
\autoref{tab:train_params} provides the hyperparameter values used in the GPI-LS MOPPO framework across all experiments conducted.

\textbf{Network architecture: }The MOPPO frmework implements the state and weight encoders as a single hidden layer with 256 units each. The actor and critic networks consist of three hidden layers of 256 units each, with ReLU activations, layer normalization, and dropout (rate 0.1) applied between layers. Optimization is performed using the Adam optimizer.

\begin{table}[t]
\caption{Hyperparameters used in the GPI-LS MOPPO framework.}
\label{tab:train_params}
\centering
\begin{tabular}{ll}
\toprule
\textit{Parameter} & \textit{Value} \\
\midrule
Learning rate & $3 \times 10^{-4}$ \\
Discount factor ($\gamma$) & 0.98 \\
Initial epsilon ($\epsilon_{init}$) & 1 \\
Final epsilon ($\epsilon_{final}$) & 0.05 \\
Epsilon decay steps & 100000 \\
Rollout buffer size & 2048 \\
\bottomrule
\end{tabular}
\end{table}

For the baseline, we used the original implementation of GPI-LS with DQN in MORL Baselines toolkit \cite{felten2023toolkit} and their original hyperparameter configurations.

All experiments were conducted on a Linux cluster node equipped with a single NVIDIA A100-SXM4 GPU (80 GB memory) and a dual-socket 48-core AMD EPYC 7642 CPU.  In our RL environment for truck driving, training our method GPI-LS MOPPO for $7.5 \times 10^{5}$ steps required approximately 30 hours, whereas the same training using the baseline method completed in roughly 35 hours.


\section{Detailed results} \label{app:results}

\subsection{Analytical solution} \label{app:anal_sol}

\begin{figure}[h!]
\centering
\includegraphics[width=0.4\textwidth]{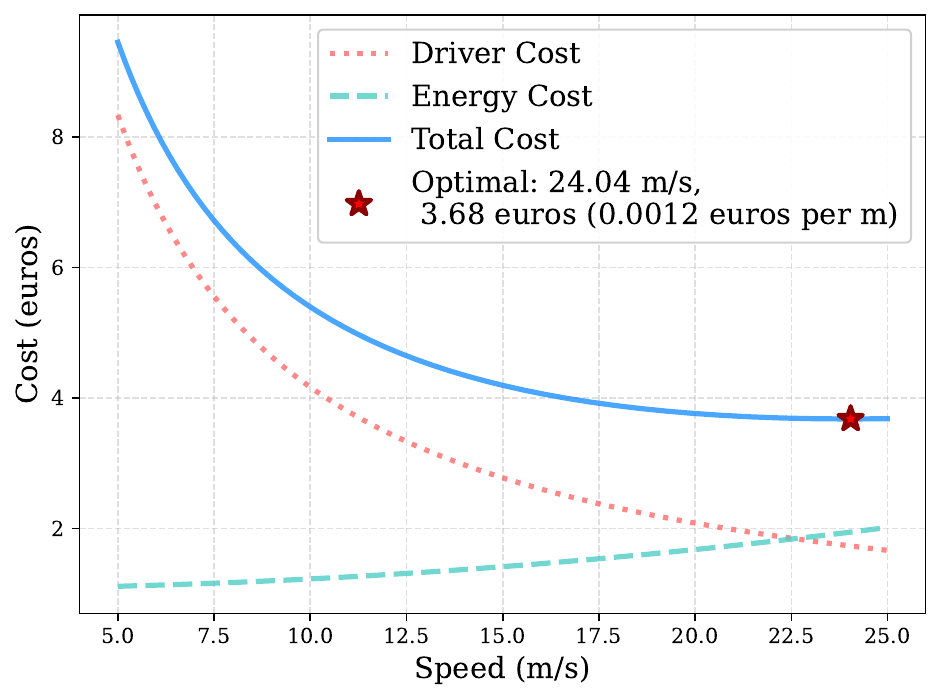}
\caption{Analytical predictions for optimal speed and cost in zero traffic situation.}
\Description{Analytical predictions for optimal speed and cost in zero traffic situation.}
\label{fig:anal_res}
\end{figure}

\autoref{fig:anal_res} shows a detailed graph of the analytical solution for a $3000 m$ road length under the constant speed assumption, which is used in Section \ref{sec:res_analy}. It plots driver cost, energy cost and total cost (which is a sum of other two costs) as a function of speed.

\subsection{Pareto fronts in different traffic settings}
\begin{figure}[h]
\centering
\includegraphics[width=0.45\textwidth]{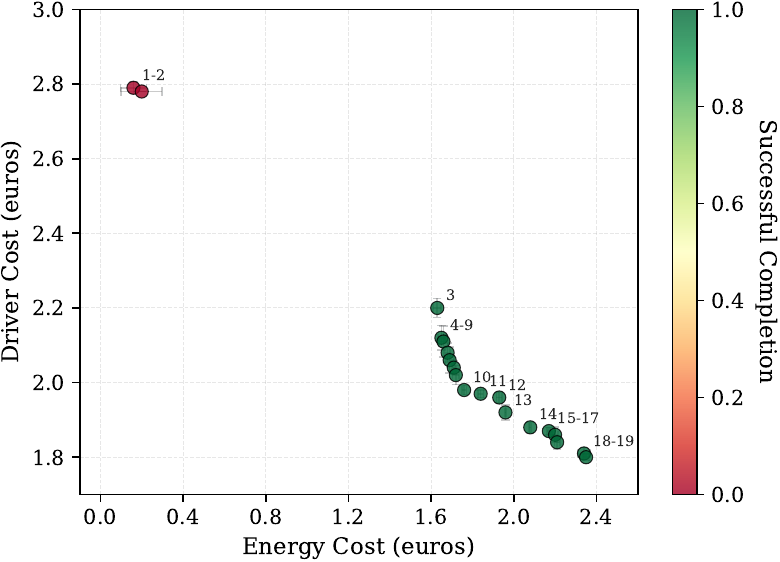}
\caption{{Pareto Front showing the trade off between driver cost and energy cost in zero traffic, along with the success rate. The annotated numbers denote the policy numbers referred in the \autoref{tab:policy_results_zero}}} 
\Description{Pareto Front showing the trade off between driver cost and energy cost in zero traffic, along with the success rate. The annotated numbers denote the policy numbers referred in the \autoref{tab:policy_results_zero}}
\label{fig:par_zero}
\end{figure}

\begin{figure}[h]
\centering
\includegraphics[width=0.45\textwidth]{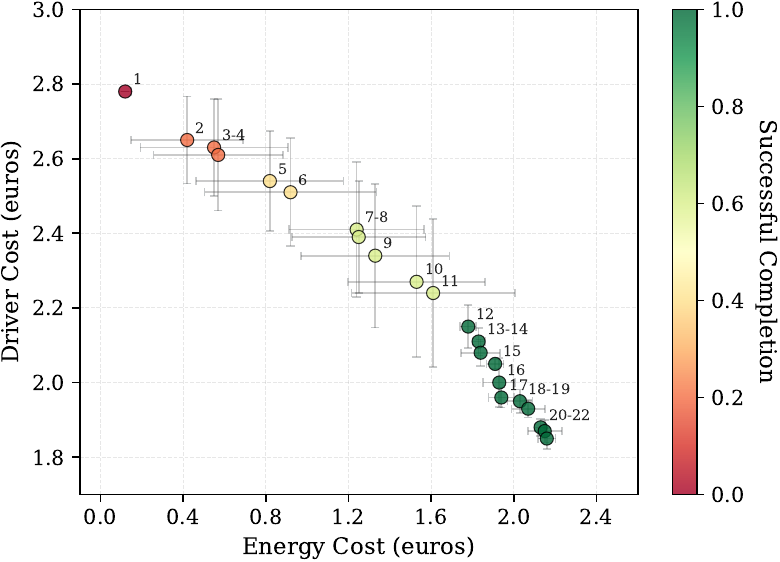}
\caption{{Pareto Front showing the trade off between driver cost and energy cost in medium traffic, along with the success rate. The annotated numbers denote the policy numbers referred in the \autoref{tab:policy_results_medium}}} 
\Description{Pareto Front showing the trade off between driver cost and energy cost in medium traffic, along with the success rate. The annotated numbers denote the policy numbers referred in the \autoref{tab:policy_results_medium}}
\label{fig:par_med}
\end{figure}

\begin{figure}[h]
\centering
\includegraphics[width=0.45\textwidth]{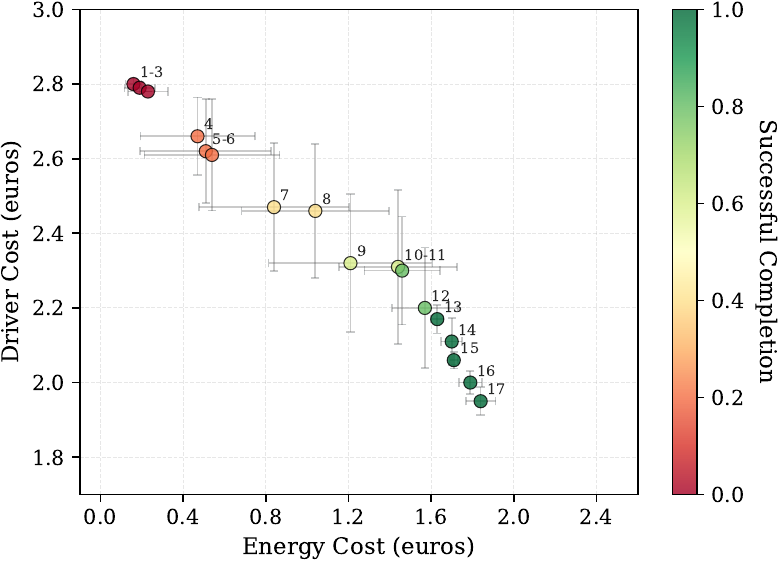}
\caption{{Pareto Front showing the trade off between driver cost and energy cost in high traffic, along with the success rate. The annotated numbers denote the policy numbers referred in the \autoref{tab:policy_results_high}}}
\Description{Pareto Front showing the trade off between driver cost and energy cost in high traffic, along with the success rate. The annotated numbers denote the policy numbers referred in the \autoref{tab:policy_results_high}}
\label{fig:par_high}
\end{figure}

Figures~\ref{fig:par_zero}, \ref{fig:par_med}, and \ref{fig:par_high} illustrate the Pareto fronts shown in \autoref{fig:pareto}, where each policy is labeled by its corresponding number. A detailed quantitative evaluation of each policy in the three traffic settings is given in Tables~\ref{tab:policy_results_zero}, \ref{tab:policy_results_medium}, and \ref{tab:policy_results_high}, which report the mean and standard error of each metric over 5 episodes.

We include feasible policies that has a collision rate of $0\%$ in the Pareto front. Therefore success rate indicates the number of episodes that reached the target within the maximum allowed steps. The max step rate corresponds to episodes that terminated after reaching the maximum number of steps without completing the target. The target distance is approximately 3000 m, with minor variations depending on the truck’s start and end positions in simulation.

The results are easily interpretable. Policies that prioritize minimizing energy cost tend to have lower average speeds, resulting in longer times to reach the target and consequently higher driver costs. Conversely, policies that prioritize minimizing driver cost tend to have higher average speeds and consequently higher energy costs. In the zero-traffic case, the best total cost achieved by successful policies is around 3.74 euros over 3 km, which is comparable to the analytical cost of 3.68 euros. The corresponding best total cost per meter is 0.0012 euros/m, which matches the analytical value. Notably, the same best total cost per meter of 0.0012 euros/m is achieved in the medium- and high-traffic cases. Although denser traffic results in a wider range of average speeds and affects the trade-off between energy and driver costs, the best-performing policies maintain a total cost per meter comparable to the analytical optimum across all three traffic conditions.

\FloatBarrier

\begin{table*}[t]
\centering
\caption{Evaluation of Pareto-optimal policies in zero traffic.}
\label{tab:policy_results_zero}
\setlength{\tabcolsep}{4pt}
\begin{tabular}{cccccccc}
\toprule
\textit{Policy} & \textit{Success} & \textit{Average} & \textit{Energy} &
\textit{Driver} & \textit{Traveled} & \textit{Total Cost} & \textit{Total Cost} \\
\textit{Number} & \textit{Rate (\%)} & \textit{Speed (m/s)} & \textit{Cost (euros)} &
\textit{Cost (euros)} & \textit{Distance (m)} & \textit{ (euros)} & \textit{per m (euros)} \\
\midrule
1 & 0 $\pm$ 0 & 1.63 $\pm$ 0.46 & 0.16 $\pm$ 0.06 & 2.79 $\pm$ 0.01 & 341 $\pm$ 100 & 2.95 $\pm$ 0.08 & 0.0086 $\pm$ 0.0025 \\
2 & 0 $\pm$ 0 & 1.84 $\pm$ 0.65 & 0.20 $\pm$ 0.10 & 2.78 $\pm$ 0.00 & 376 $\pm$ 130 & 2.98 $\pm$ 0.10 & 0.0079 $\pm$ 0.0027 \\
3 & 100 $\pm$ 0 & 19.03 $\pm$ 0.26 & 1.63 $\pm$ 0.02 & 2.20 $\pm$ 0.03 & 3012 $\pm$ 10 & 3.83 $\pm$ 0.01 & 0.0013 $\pm$ 0.0000 \\
4 & 100 $\pm$ 0 & 19.59 $\pm$ 0.30 & 1.65 $\pm$ 0.02 & 2.12 $\pm$ 0.03 & 2989 $\pm$ 4 & 3.77 $\pm$ 0.02 & 0.0013 $\pm$ 0.0000 \\
5 & 100 $\pm$ 0 & 19.73 $\pm$ 0.38 & 1.66 $\pm$ 0.02 & 2.11 $\pm$ 0.04 & 2993 $\pm$ 4 & 3.77 $\pm$ 0.02 & 0.0013 $\pm$ 0.0000 \\
6 & 100 $\pm$ 0 & 20.03 $\pm$ 0.23 & 1.68 $\pm$ 0.01 & 2.08 $\pm$ 0.03 & 2995 $\pm$ 9 & 3.76 $\pm$ 0.02 & 0.0013 $\pm$ 0.0000 \\
7 & 100 $\pm$ 0 & 20.21 $\pm$ 0.34 & 1.69 $\pm$ 0.02 & 2.06 $\pm$ 0.04 & 2986 $\pm$ 4 & 3.75 $\pm$ 0.02 & 0.0013 $\pm$ 0.0000 \\
8 & 100 $\pm$ 0 & 20.43 $\pm$ 0.14 & 1.71 $\pm$ 0.01 & 2.04 $\pm$ 0.02 & 3006 $\pm$ 11 & 3.75 $\pm$ 0.02 & 0.0012 $\pm$ 0.0000 \\
9 & 100 $\pm$ 0 & 20.66 $\pm$ 0.24 & 1.72 $\pm$ 0.02 & 2.02 $\pm$ 0.03 & 3006 $\pm$ 13 & 3.74 $\pm$ 0.02 & 0.0012 $\pm$ 0.0000 \\
10 & 100 $\pm$ 0 & 21.15 $\pm$ 0.14 & 1.76 $\pm$ 0.01 & 1.98 $\pm$ 0.01 & 3011 $\pm$ 7 & 3.74 $\pm$ 0.01 & 0.0012 $\pm$ 0.0000 \\
11 & 100 $\pm$ 0 & 21.21 $\pm$ 0.15 & 1.84 $\pm$ 0.01 & 1.97 $\pm$ 0.01 & 3018 $\pm$ 10 & 3.81 $\pm$ 0.01 & 0.0013 $\pm$ 0.0000 \\
12 & 100 $\pm$ 0 & 21.31 $\pm$ 0.18 & 1.93 $\pm$ 0.01 & 1.96 $\pm$ 0.01 & 3020 $\pm$ 13 & 3.89 $\pm$ 0.02 & 0.0013 $\pm$ 0.0000 \\
13 & 100 $\pm$ 0 & 21.64 $\pm$ 0.18 & 1.96 $\pm$ 0.00 & 1.92 $\pm$ 0.02 & 3008 $\pm$ 10 & 3.88 $\pm$ 0.02 & 0.0013 $\pm$ 0.0000 \\
14 & 100 $\pm$ 0 & 21.91 $\pm$ 0.21 & 2.08 $\pm$ 0.01 & 1.88 $\pm$ 0.01 & 2997 $\pm$ 7 & 3.96 $\pm$ 0.01 & 0.0013 $\pm$ 0.0000 \\
15 & 100 $\pm$ 0 & 22.02 $\pm$ 0.15 & 2.17 $\pm$ 0.01 & 1.87 $\pm$ 0.01 & 3014 $\pm$ 11 & 4.04 $\pm$ 0.02 & 0.0013 $\pm$ 0.0000 \\
16 & 100 $\pm$ 0 & 22.14 $\pm$ 0.20 & 2.20 $\pm$ 0.02 & 1.86 $\pm$ 0.02 & 3016 $\pm$ 14 & 4.06 $\pm$ 0.04 & 0.0013 $\pm$ 0.0000 \\
17 & 100 $\pm$ 0 & 22.19 $\pm$ 0.22 & 2.21 $\pm$ 0.01 & 1.84 $\pm$ 0.02 & 2991 $\pm$ 6 & 4.05 $\pm$ 0.03 & 0.0014 $\pm$ 0.0000 \\
18 & 100 $\pm$ 0 & 22.96 $\pm$ 0.20 & 2.34 $\pm$ 0.03 & 1.81 $\pm$ 0.02 & 2988 $\pm$ 3 & 4.15 $\pm$ 0.04 & 0.0014 $\pm$ 0.0000 \\
19 & 100 $\pm$ 0 & 23.03 $\pm$ 0.09 & 2.35 $\pm$ 0.01 & 1.80 $\pm$ 0.01 & 2983 $\pm$ 2 & 4.15 $\pm$ 0.02 & 0.0014 $\pm$ 0.0000 \\
\bottomrule
\end{tabular}
\end{table*}

\begin{table*}[t]
\centering
\caption{Evaluation of Pareto-optimal policies in medium traffic.}
\label{tab:policy_results_medium}
\setlength{\tabcolsep}{4pt}
\begin{tabular}{cccccccc}
\toprule
\textit{Policy} & \textit{Success} & \textit{Average} & \textit{Energy} &
\textit{Driver} & \textit{Traveled} & \textit{Total Cost} & \textit{Total Cost} \\
\textit{Number} & \textit{Rate (\%)} & \textit{Speed (m/s)} & \textit{Cost (euros)} &
\textit{Cost (euros)} & \textit{Distance (m)} & \textit{ (euros)} & \textit{per m (euros)} \\
\midrule
1 & 0 $\pm$ 0 & 1.41 $\pm$ 0.27 & 0.12 $\pm$ 0.03 & 2.78 $\pm$ 0.00 & 371 $\pm$ 48 & 2.90 $\pm$ 0.03 & 0.0078 $\pm$ 0.0010 \\
2 & 20 $\pm$ 18 & 5.03 $\pm$ 3.22 & 0.42 $\pm$ 0.27 & 2.65 $\pm$ 0.12 & 990 $\pm$ 477 & 3.07 $\pm$ 0.16 & 0.0031 $\pm$ 0.0015 \\
3 & 20 $\pm$ 18 & 5.35 $\pm$ 3.26 & 0.55 $\pm$ 0.36 & 2.63 $\pm$ 0.13 & 930 $\pm$ 496 & 3.18 $\pm$ 0.23 & 0.0034 $\pm$ 0.0018 \\
4 & 20 $\pm$ 18 & 6.40 $\pm$ 3.35 & 0.57 $\pm$ 0.31 & 2.61 $\pm$ 0.15 & 999 $\pm$ 502 & 3.18 $\pm$ 0.17 & 0.0032 $\pm$ 0.0016 \\
5 & 40 $\pm$ 22 & 8.66 $\pm$ 3.67 & 0.82 $\pm$ 0.36 & 2.54 $\pm$ 0.13 & 1373 $\pm$ 577 & 3.36 $\pm$ 0.23 & 0.0024 $\pm$ 0.0010 \\
6 & 40 $\pm$ 22 & 8.76 $\pm$ 3.81 & 0.92 $\pm$ 0.42 & 2.51 $\pm$ 0.14 & 1499 $\pm$ 600 & 3.43 $\pm$ 0.27 & 0.0023 $\pm$ 0.0009 \\
7 & 60 $\pm$ 22 & 13.46 $\pm$ 3.49 & 1.24 $\pm$ 0.33 & 2.41 $\pm$ 0.18 & 2207 $\pm$ 453 & 3.65 $\pm$ 0.16 & 0.0017 $\pm$ 0.0003 \\
8 & 60 $\pm$ 22 & 13.24 $\pm$ 3.39 & 1.25 $\pm$ 0.32 & 2.39 $\pm$ 0.15 & 2218 $\pm$ 520 & 3.64 $\pm$ 0.20 & 0.0016 $\pm$ 0.0004 \\
9 & 60 $\pm$ 22 & 13.83 $\pm$ 3.73 & 1.33 $\pm$ 0.36 & 2.34 $\pm$ 0.19 & 1994 $\pm$ 540 & 3.67 $\pm$ 0.17 & 0.0018 $\pm$ 0.0005 \\
10 & 60 $\pm$ 22 & 15.46 $\pm$ 3.22 & 1.53 $\pm$ 0.33 & 2.27 $\pm$ 0.20 & 2437 $\pm$ 358 & 3.80 $\pm$ 0.13 & 0.0016 $\pm$ 0.0002 \\
11 & 60 $\pm$ 22 & 15.32 $\pm$ 3.54 & 1.61 $\pm$ 0.39 & 2.24 $\pm$ 0.20 & 2240 $\pm$ 483 & 3.85 $\pm$ 0.20 & 0.0017 $\pm$ 0.0004 \\
12 & 100 $\pm$ 0 & 19.43 $\pm$ 0.43 & 1.78 $\pm$ 0.04 & 2.15 $\pm$ 0.06 & 3168 $\pm$ 16 & 3.93 $\pm$ 0.07 & 0.0012 $\pm$ 0.0000 \\
13 & 100 $\pm$ 0 & 19.74 $\pm$ 0.20 & 1.83 $\pm$ 0.03 & 2.11 $\pm$ 0.04 & 2916 $\pm$ 77 & 3.94 $\pm$ 0.06 & 0.0014 $\pm$ 0.0000 \\
14 & 100 $\pm$ 0 & 19.92 $\pm$ 0.39 & 1.84 $\pm$ 0.09 & 2.08 $\pm$ 0.04 & 3135 $\pm$ 16 & 3.92 $\pm$ 0.07 & 0.0013 $\pm$ 0.0000 \\
15 & 100 $\pm$ 0 & 20.03 $\pm$ 0.10 & 1.91 $\pm$ 0.04 & 2.05 $\pm$ 0.01 & 3071 $\pm$ 35 & 3.96 $\pm$ 0.05 & 0.0013 $\pm$ 0.0000 \\
16 & 100 $\pm$ 0 & 20.50 $\pm$ 0.62 & 1.93 $\pm$ 0.08 & 2.00 $\pm$ 0.07 & 3033 $\pm$ 79 & 3.93 $\pm$ 0.04 & 0.0013 $\pm$ 0.0000 \\
17 & 100 $\pm$ 0 & 20.81 $\pm$ 0.25 & 1.94 $\pm$ 0.06 & 1.96 $\pm$ 0.03 & 3033 $\pm$ 72 & 3.90 $\pm$ 0.05 & 0.0013 $\pm$ 0.0000 \\
18 & 100 $\pm$ 0 & 20.93 $\pm$ 0.38 & 2.03 $\pm$ 0.06 & 1.95 $\pm$ 0.03 & 3036 $\pm$ 70 & 3.98 $\pm$ 0.03 & 0.0013 $\pm$ 0.0000 \\
19 & 100 $\pm$ 0 & 21.14 $\pm$ 0.26 & 2.07 $\pm$ 0.08 & 1.93 $\pm$ 0.02 & 3113 $\pm$ 13 & 4.00 $\pm$ 0.07 & 0.0013 $\pm$ 0.0000 \\
20 & 100 $\pm$ 0 & 21.61 $\pm$ 0.28 & 2.13 $\pm$ 0.03 & 1.88 $\pm$ 0.02 & 2960 $\pm$ 83 & 4.01 $\pm$ 0.04 & 0.0014 $\pm$ 0.0000 \\
21 & 100 $\pm$ 0 & 21.76 $\pm$ 0.33 & 2.15 $\pm$ 0.08 & 1.87 $\pm$ 0.03 & 3076 $\pm$ 19 & 4.02 $\pm$ 0.06 & 0.0013 $\pm$ 0.0000 \\
22 & 100 $\pm$ 0 & 22.09 $\pm$ 0.28 & 2.16 $\pm$ 0.04 & 1.85 $\pm$ 0.03 & 3120 $\pm$ 20 & 4.01 $\pm$ 0.04 & 0.0013 $\pm$ 0.0000 \\
\bottomrule
\end{tabular}
\end{table*}
   
\begin{table*}[t]
\centering
\caption{Evaluation of Pareto-optimal policies in high traffic.}
\label{tab:policy_results_high}
\setlength{\tabcolsep}{4pt}
\begin{tabular}{cccccccc}
\toprule
\textit{Policy} & \textit{Success} & \textit{Average} & \textit{Energy} &
\textit{Driver} & \textit{Traveled} & \textit{Total Cost} & \textit{Total Cost} \\
\textit{Number} & \textit{Rate (\%)} & \textit{Speed (m/s)} & \textit{Cost (euros)} &
\textit{Cost (euros)} & \textit{Distance (m)} & \textit{ (euros)} & \textit{per m (euros)} \\
\midrule
1 & 0 $\pm$ 0 & 1.65 $\pm$ 0.32 & 0.16 $\pm$ 0.04 & 2.80 $\pm$ 0.01 & 252 $\pm$ 89 & 2.96 $\pm$ 0.05 & 0.0118 $\pm$ 0.0041 \\
2 & 0 $\pm$ 0 & 1.96 $\pm$ 0.58 & 0.19 $\pm$ 0.07 & 2.79 $\pm$ 0.01 & 389 $\pm$ 207 & 2.98 $\pm$ 0.09 & 0.0077 $\pm$ 0.0041 \\
3 & 0 $\pm$ 0 & 2.13 $\pm$ 0.71 & 0.23 $\pm$ 0.10 & 2.78 $\pm$ 0.00 & 203 $\pm$ 143 & 3.01 $\pm$ 0.10 & 0.0148 $\pm$ 0.0105 \\
4 & 20 $\pm$ 18 & 4.98 $\pm$ 2.97 & 0.47 $\pm$ 0.28 & 2.66 $\pm$ 0.10 & 906 $\pm$ 512 & 3.13 $\pm$ 0.17 & 0.0035 $\pm$ 0.0020 \\
5 & 20 $\pm$ 18 & 5.38 $\pm$ 3.26 & 0.51 $\pm$ 0.32 & 2.62 $\pm$ 0.14 & 735 $\pm$ 496 & 3.13 $\pm$ 0.18 & 0.0043 $\pm$ 0.0029 \\
6 & 20 $\pm$ 18 & 5.50 $\pm$ 3.36 & 0.54 $\pm$ 0.33 & 2.61 $\pm$ 0.15 & 812 $\pm$ 521 & 3.15 $\pm$ 0.18 & 0.0039 $\pm$ 0.0025 \\
7 & 40 $\pm$ 22 & 9.29 $\pm$ 3.96 & 0.84 $\pm$ 0.36 & 2.47 $\pm$ 0.17 & 1319 $\pm$ 627 & 3.31 $\pm$ 0.20 & 0.0025 $\pm$ 0.0012 \\
8 & 40 $\pm$ 22 & 10.77 $\pm$ 3.86 & 1.04 $\pm$ 0.36 & 2.46 $\pm$ 0.18 & 1811 $\pm$ 579 & 3.50 $\pm$ 0.23 & 0.0019 $\pm$ 0.0006 \\
9 & 60 $\pm$ 22 & 12.74 $\pm$ 4.11 & 1.21 $\pm$ 0.40 & 2.32 $\pm$ 0.18 & 1872 $\pm$ 575 & 3.53 $\pm$ 0.21 & 0.0019 $\pm$ 0.0006 \\
10 & 60 $\pm$ 22 & 15.42 $\pm$ 3.27 & 1.44 $\pm$ 0.29 & 2.31 $\pm$ 0.21 & 2458 $\pm$ 404 & 3.75 $\pm$ 0.18 & 0.0015 $\pm$ 0.0003 \\
11 & 80 $\pm$ 18 & 16.14 $\pm$ 2.31 & 1.46 $\pm$ 0.18 & 2.30 $\pm$ 0.14 & 2569 $\pm$ 361 & 3.76 $\pm$ 0.05 & 0.0015 $\pm$ 0.0002 \\
12 & 80 $\pm$ 18 & 17.17 $\pm$ 2.37 & 1.57 $\pm$ 0.16 & 2.20 $\pm$ 0.16 & 2789 $\pm$ 239 & 3.77 $\pm$ 0.01 & 0.0014 $\pm$ 0.0001 \\
13 & 100 $\pm$ 0 & 18.44 $\pm$ 0.33 & 1.63 $\pm$ 0.03 & 2.17 $\pm$ 0.04 & 3042 $\pm$ 24 & 3.80 $\pm$ 0.02 & 0.0012 $\pm$ 0.0000 \\
14 & 100 $\pm$ 0 & 19.04 $\pm$ 0.57 & 1.70 $\pm$ 0.05 & 2.11 $\pm$ 0.06 & 3014 $\pm$ 28 & 3.81 $\pm$ 0.03 & 0.0013 $\pm$ 0.0000 \\
15 & 100 $\pm$ 0 & 19.44 $\pm$ 0.23 & 1.71 $\pm$ 0.02 & 2.06 $\pm$ 0.02 & 2951 $\pm$ 63 & 3.77 $\pm$ 0.02 & 0.0013 $\pm$ 0.0000 \\
16 & 100 $\pm$ 0 & 20.00 $\pm$ 0.31 & 1.79 $\pm$ 0.06 & 2.00 $\pm$ 0.03 & 2972 $\pm$ 72 & 3.79 $\pm$ 0.04 & 0.0013 $\pm$ 0.0000 \\
17 & 100 $\pm$ 0 & 20.49 $\pm$ 0.41 & 1.84 $\pm$ 0.07 & 1.95 $\pm$ 0.04 & 3079 $\pm$ 22 & 3.79 $\pm$ 0.04 & 0.0012 $\pm$ 0.0000 \\
\bottomrule
\end{tabular}
\end{table*}

\subsection{Baseline comparison}

\paragraph{Execution time}
We observed that our GPI-LS MOPPO framework is computationally more efficient than GPI-LS based on DQN \cite{sample_eff} as implemented in \cite{felten2023toolkit}. In our RL environment for truck driving, training the latter for $7.5 \times 10^{5}$ steps required approximately 35 hours, whereas the same training using our framework completed in roughly 30 hours. Experiments were conducted on a Linux cluster node equipped with a single NVIDIA A100-SXM4 GPU (80 GB memory) and a dual-socket  48-core AMD EPYC 7642 CPU.

\begin{table*}[t]
\caption{Evaluation metrics over the training checkpoints for the baseline (GPI-LS with DQN) and our method (GPI-LS MOPPO).}
\label{tab:eval_comp}
\centering
\begin{tabular}{rcccc}
\toprule
\textit{Training Steps} &
\multicolumn{2}{c}{\textit{Expected Utility}} &
\multicolumn{2}{c}{\textit{Mean Utility}} \\
\cmidrule(lr){2-3}
\cmidrule(lr){4-5}
& \textit{Baseline} & \textit{Ours}
& \textit{Baseline} & \textit{Ours} \\
\midrule
300k & -0.1704 $\pm$ 0.0415 & 0.3588 $\pm$ 0.0216
     & -85.5685 $\pm$ 31.3080 & -1.7035 $\pm$ 1.4485 \\
350k & -0.1455 $\pm$ 0.0332 & 0.3270 $\pm$ 0.0183
     & -19.8703 $\pm$ 2.9108 & -0.8235 $\pm$ 0.6110 \\
400k & -0.1675 $\pm$ 0.0238 & 0.1883 $\pm$ 0.1655
     & -29.0711 $\pm$ 27.2000 & -0.8603 $\pm$ 0.1989 \\
450k & -0.1355 $\pm$ 0.0068 & 0.3311 $\pm$ 0.0255
     & -25.6834 $\pm$ 21.3722 & -0.2739 $\pm$ 0.2631 \\
500k & -0.1593 $\pm$ 0.0353 & 0.3419 $\pm$ 0.0173
     & -22.7032 $\pm$ 20.9624 & -0.3035 $\pm$ 0.3839 \\
\bottomrule
\end{tabular}
\end{table*}

\paragraph{Evaluation metrics} \autoref{tab:eval_comp} reports two evaluation metrics: Expected Utility and Mean Utility, evaluated over a set of 50 sampled evaluation weights, during different training checkpoints. 

\paragraph{Expected Utility}
Expected Utility evaluates the expected maximum scalarized return over a distribution of weights, computed from a set of value vectors ${\mathcal{V}} = \{\mathbf{v}^\pi\}$ corresponding to learned policies.
\begin{equation}
\operatorname{EU}=\mathbb{E}_{\mathbf{w} \sim \mathcal{W}}\left[\max_{\mathbf{v}^\pi \in {\mathcal{V}}} \mathbf{w}^\top \mathbf{v}^\pi \right]
\end{equation}

\paragraph{Mean Utility}
Mean utility evaluates the average scalarized return $\mathbf{v}_i$ obtained by the policy conditioned by each test preference $\mathbf{w}_i$:
\begin{equation}
\hat{J} = \frac{1}{K} \sum_{i=1}^{K} \mathbf{w}_i^\top \mathbf{v}_i
\end{equation}
This measures the average performance of the trained model across the test preferences.

The results in \autoref{tab:eval_comp} indicates better Expected Utility and Mean Utility for our method compared to the baseline.


\end{document}